\documentclass[a4paper,fleqn]{cas-dc}

\usepackage[authoryear,longnamesfirst]{natbib}
\usepackage{graphicx}%
\usepackage{subcaption}%
\usepackage{multirow}%
\usepackage{amsmath,amssymb,amsfonts}%
\usepackage{algorithm}%
\usepackage{algorithmicx}%
\usepackage{algpseudocode}%
\usepackage{listings}%
\usepackage{url}%
\usepackage{tabularx} %

\def\tsc#1{\csdef{#1}{\textsc{\lowercase{#1}}\xspace}}
\tsc{WGM}
\tsc{QE}
\tsc{EP}
\tsc{PMS}
\tsc{BEC}
\tsc{DE}

\begin{document}
\let\WriteBookmarks\relax
\def\floatpagepagefraction{1}
\def\textpagefraction{.001}

\shorttitle{Explainability Analysis - Large Language Model (LLM) Approach}

\shortauthors{Uddin et~al.}


\title [mode = title]{ExplainableDetector: Exploring Transformer-based Language Modeling Approach for SMS Spam Detection with Explainability Analysis} 



%
\author[1]{Mohammad Amaz Uddin}[]



\ead{amazuddin722@gmail.com}


\credit{Writing – original draft, Methodology, Data curation, Implementation, Experimental analysis, Conceptualization}

\affiliation[1]{organization={Department of Computer Science and Engineering, BGC Trust University Bangladesh},
    city={Chittagong},
    postcode={4381}, 
    country={Bangladesh}}

\author[2]{Muhammad Nazrul Islam}[]

\credit{Writing – review \& editing}

\affiliation[2]{organization={Department of Computer Science and Engineering, Military Institute of Science and Technology},
    city={Dhaka},
    postcode={1216}, 
    country={Bangladesh}}

\author[3]{Leandros Maglaras}[]

\credit{Writing – review \& editing}

\affiliation[3]{organization={School of Computing, Edinburgh Napier University},
    country={UK}}

\author[4]{Helge Janicke}[]

\credit{Writing – review \& editing}

\author[4]{Iqbal H. Sarker}[
                        orcid=https://orcid.org/0000-0003-1740-5517]
                        

\cormark[1]
\ead{m.sarker@ecu.edu.au}

\credit{Writing – review \& editing, Conceptualization and Supervision}

\affiliation[4]{organization={Centre for Securing Digital Futures, Edith Cowan University},
    city={Perth},
    postcode={WA-6027}, 
    country={Australia}}

\cortext[cor1]{Corresponding author}


\begin{abstract}
SMS, or short messaging service, is a widely used and cost-effective communication medium that has sadly turned into a haven for unwanted messages, commonly known as SMS spam. With the rapid adoption of smartphones and Internet connectivity, SMS spam has emerged as a prevalent threat. Spammers have taken notice of the significance of SMS for mobile phone users. Consequently, with the emergence of new cybersecurity threats, the number of SMS spam has expanded significantly in recent years. The unstructured format of SMS data creates significant challenges for SMS spam detection, making it more difficult to successfully fight spam attacks in the cybersecurity domain. In this work, we employ optimized and fine-tuned transformer-based Large Language Models (LLMs) to solve the problem of spam message detection. We use a benchmark SMS spam dataset for this spam detection and utilize several preprocessing techniques to get clean and noise-free data and solve the class imbalance problem using the text augmentation technique. The overall experiment showed that our optimized fine-tuned BERT (Bidirectional Encoder Representations from Transformers) variant model RoBERTa obtained high accuracy with 99.84\%. We also work with Explainable Artificial Intelligence (XAI) techniques to calculate the positive and negative coefficient scores which explore and explain the fine-tuned model transparency in this text-based spam SMS detection task. In addition, traditional Machine Learning (ML) models were also examined to compare their performance with the transformer-based models. This analysis describes how LLMs can make a good impact on complex textual-based spam data in the cybersecurity field.

\end{abstract}




\maketitle

\section{Introduction}

In recent years, SMS has become the primary communication tool as mobile phones and networks have proliferated \cite{liu2021spam}. It is quite popular since it's quick and convenient, and it responds to messages faster than emails \cite{oswald2022spotspam}. Due to its convenience, accessibility, and broad use in our increasingly interconnected society, SMS is now the most popular form of communication for both individuals and organizations. Day by day, SMS has developed into a powerful tool for global communication due to its several benefits, such as simplicity of use, cost-effectiveness, wide reach, and more. Additionally, SMS promotes customer relationships, provides company updates, and provides discreet individualized communication, all of which contribute to real-time audience engagement. It is effectively used in various industries, including marketing, retail, healthcare, and finance. However, because of these benefits, an increasing number of mobile users are utilizing SMS facilities, which encourages spammers to generate spam SMS attacks aimed at harming users \cite{hossain2022detecting}. Spammers primarily gather phone numbers via a variety of methods and then utilize automated software services to deliver large volumes of messages to these numbers. Phishing URLs, fraudulent schemes, and other fraudulent content are promoted in these bulk messages. Financial gain is their primary objective, which is attained by purchasing counterfeit items, obtaining personal information, or clicking on malicious links that download malware \cite{wei2020lightweight}. Although spam detection is a challenging task in the cybersecurity field, significant enhancements to these efforts can be achieved by leveraging Artificial Intelligence (AI) methods, including Machine Learning, Deep Learning, Hybrid Learning, and Transfer Learning, which have proven effective in enhancing spam detection efforts \cite{sarker2023machine}. Furthermore, due to rapid technological advancements, novel AI-based methods are being developed daily within the cybersecurity field \cite{amaz2024explainable}.

LLMs research, supported by both industry and academia, has been essential in the recent advances in natural language interpretation \cite{zhao2023survey}. The field of artificial intelligence research and application could undergo a revolutionary shift because of these models. These language models trained on expansive datasets, they excel in capturing nuanced linguistic patterns. Utilizing the power of LLMs, companies like Google, Microsoft, Amazon, and OpenAI are developing different AI tools that are skilled at understanding and providing natural language responses to user's requests. Notably, OpenAI's ChatGPT and Bard exemplify the transformative power of LLMs, captivating widespread attention for their ability to generate and understand human-like text \cite{yao2023survey}. For natural language processing applications, transformer-based models such as GPT (Generative Pre-trained Transformer) and Google's BERT (Bidirectional Encoder Representations from Transformers) are referred to as LLMs because of their capacity to understand and generate text that is similar to that of a human. These transformer-based models are useful for a variety of tasks in NLP research and applications because of their versatility in handling translation, generation, and language comprehension.

In the field of NLP, a Transformer \cite{vaswani2017attention} refers to a deep learning model architecture that is becoming more popular in text classification, and translation tasks. Its creative architecture makes use of attention mechanisms to reliably encode input data into representations, providing unmatched performance across a range of natural language processing applications \cite{han2021transformer}. Transformer-based models have brought in a new era of deep learning capabilities, demonstrating their transformational impact on the discipline in areas ranging from spam categorization to text comprehension and interpretation \cite{jamal2023improved}. Recently, improved transformer-based models, such as BERT, were presented to address several NLP problems. The BERT-language representation model is a transformer-based neural network architecture that has been pre-trained. It is extremely effective and useful for a variety of text-related applications, such as spam and phishing detection \cite{koroteev2021bert}. By utilizing bidirectional attention mechanisms, BERT can comprehend the complex relationships between words and phrases within a sentence by grasping contextual word embeddings \cite{mohammed2021survey}. The ability of BERT to take into account context from both ends (left and right) of a sentence sets it apart from previous models that were restricted to unidirectional processing \cite{devlin2018bert}. This bidirectional approach gives BERT a better understanding of the language nuances and relationships present in the text. Using a Masked Language Model (MLM) approach, BERT trains the model to infer the original words from context by substituting specific words in a sentence with a unique token ([MASK]). Furthermore, BERT has other variations, including BERT-base, BERT-large \cite{anil2021large}, ALBERT \cite{lan2019albert}, RoBERTa \cite{liu2019roberta}, DistilBERT \cite{sanh2019distilbert}, and TinyBERT \cite{jiao2019tinybert} among others. These variations vary in the number of transformer layers and parameters, and each one provides unique benefits and enhancements for tasks involving natural language processing.

BERT is useful in a variety of NLP tasks, but because of its complexity, it is sometimes seen as a "black box" making it difficult to comprehend how it makes predictions. XAI techniques have become crucial tools for clarifying complex procedures \cite{sarker2024ai, xu2019explainable,linardatos2020explainable}. Through exploring the model's inner workings, these methods provide insightful explanations for why particular predictions are made. By using XAI methods, researchers can clarify the keywords or characteristics that the model identified as most important for certain activities, which improves clarity and transparency. In this research work, our main aim is to clarify the transparency of the model in the SMS spam classification task. In summary, our major contributions to the work can be outlined as follows:

\begin{itemize}
\item We explore transformer-based LLMs for SMS spam detection. To make the model more efficient and impactful on textual-based spam data, we optimized and fine-tuned the DistilBERT and RoBERTa models.

\item We calculate the positive coefficients and negative coefficients of the text using XAI techniques to find out the spam and ham messages. This also helps to explain and investigate the interpretability of the fine-tuned model.

\item We visualize the experimental performance comparison between ML models and LLMs using balanced and imbalanced datasets. To address the class imbalance problem in the data, we employ a text augmentation technique. This also analyzes the impact of imbalanced and balanced data.
\end{itemize}

The remainder of the paper is organized as follows: Section 2 reviews related works. Section 3 presents a comprehensive explanation of the overall methodology. Section 4 conducts an in-depth analysis of the experimental results. Section 5 examines the transparency of the model using XAI approaches. Furthermore, Section 6 provides a detailed discussion of the findings and results. Finally, Section 7 concludes and discusses future work.

\section{Literature Review}

In the rapidly evolving cybersecurity landscape, combating spam SMS messages has become a major opportunity and challenge due to its sophisticated and increasingly significant nature in today's increasingly interconnected world. Indeed, researchers have created several solutions for dealing with spam SMS by utilizing cutting-edge methods from machine learning, deep learning, and hybrid models. These initiatives highlight how constantly evolving the sector is, with new techniques like NLP, anomaly detection, pattern recognition, and ensemble methods being employed to identify and counter cyber threats as they arise.

\subsection{Machine Learning Techniques}\label{subsec1}

To prevent security threats, maintain customer engagement, safeguard their reputation, enhance operational efficiency, and comply with regulations, organizations must quickly detect spam attempts. By implementing efficient spam detection systems, businesses can preserve customer confidence and secure communication channels. Many methods utilized in the field of SMS spam classification tasks in the past have contributed to achieving these objectives. \cite{abid2022spam} presented a method that utilizes supervised machine learning techniques, including TF-IDF and bag-of-words for feature extraction, to classify spam and ham SMS. To address data imbalance, oversampling and undersampling techniques were implemented. Among several machine learning classifiers tested, random forest achieved a remarkable 99\% accuracy in distinguishing between spam and ham SMS. The study \cite{mishra2023dsmishsms} represented a two-phase smishing (SMS phishing) detection model: Domain Checking Phase verifies URL (Uniform Resource Locator) authenticity, while the SMS Classification Phase extracts text features. By using the Backpropagation Algorithm, the system outperforms traditional classifiers and classifies messages with an accuracy of 97.93\% which demonstrates high efficiency in detecting smishing messages. \cite{sonowal2020detecting} showed comprehensive analysis which underscores the effectiveness of feature selection techniques in enhancing Smishing detection accuracy. This work investigates feature ranking using four correlation algorithms such as Pearson rank correlation, Spearman’s rank correlation, Kendall rank correlation, and Point biserial rank correlation alongside a machine learning approach for Smishing detection, revealing AdaBoostclassifier as the most accurate. Furthermore, with the ranking algorithm that is Kendall, rank correlation offered better accuracy than the other correlation algorithms that is 98.40\%. A machine learning model using the Naïve Bayes methodology presented for SMS spam detection and classification in ref. \cite{asaju2021short}. Using string-to-word vector feature extraction, the model efficiently handles data from the UCL repository, demonstrating its high efficacy in detecting new SMS spam with impressive classification accuracies of 99.42\% for correctly classified samples and 0.57\% for incorrectly classified ones. Several researchers have proposed a naive Bayes approach \cite{Mishra2020smishing, joo2017s} for classifying text messages as spam or ham.

\subsection{Deep Learning Techniques}\label{subsec2}
In the cybersecurity field, deep learning algorithms have become indispensable, due to their ability to process complex data structures allowing them to detect subtle patterns and anomalies. Through the utilization of deep learning, cybersecurity experts can strengthen defense systems against constantly changing dangers, guaranteeing strong safeguarding for critical data and digital assets. \cite{xia2020discrete} presented a unique method that leverages word order information to address the problem of lower frequency concerns in SMS spam detection: discrete Hidden Markov Models (HMM). The suggested HMM approach showed strong performance equivalent to deep learning techniques like CNN (Convolutional Neural Network) and LSTM (Long Short-Term Memory) models through experiments on the UCI SMS spam dataset and a Chinese SMS spam dataset. It achieved 95.9\% accuracy on the UCI dataset and demonstrates language-agnostic effectiveness in identifying spam SMS. \cite{ghourabi2020hybrid} suggested using a hybrid deep learning model to identify spam SMS messages in an English and Arabic-supporting mixed mobile environment. CNN and LSTM, two deep learning techniques, are combined in this detection model. Other popular machine learning algorithms were also used to assess the overall procedure, and the results showed that the CNN\-LSTM model that was suggested performed better than the others, achieving an accuracy of 98.37\%. The study \cite{roy2020deep} employed two deep learning techniques together: CNN and LSTM to classify Spam and Not-Spam text messages. They used a spam dataset where 747 Spam and 4,827 Not-Spam text messages exist. To evaluate the performance of the proposed method, the hybrid model was compared with the traditional machine learning models and achieved a comparatively high accuracy of 99.44\%. 
In ref. \cite{Abayomi2022deep} a deep learning model built on top of BiLSTM is presented, and its effectiveness is evaluated against the latest machine learning techniques on two datasets: the UCI SMS dataset and a recently gathered dataset named ExAIS\_SMS. The BiLSTM model outperformed certain ML classifiers, showing a moderate improvement in accuracy over some, with 93.4\% accuracy for the ExAIS\_SMS dataset and 98.6\% accuracy for the UCI dataset, after evaluation using many metrics including TP, FP, F-measure, recall, precision, and overall accuracy. Further supporting the superiority of the BiLSTM model in spam identification, the comparison also showed significant accuracy differences among ML classifiers.

\subsection{Transformer model-based Techniques}\label{subsec3}

Transformer-based models like BERT and GPT  are very useful for processing and comprehending textual data, and they can also be applied to improve cybersecurity defenses and address the ever-changing cyber threat landscape. The study \cite{sahmoud2022spam} demonstrated how well BERT performs comprehensive context analysis to enhance spam detection and produce better classification results. Using a pre-trained BERT model, a proposed high-performance spam detector obtains impressive accuracy rates (from 97.83\% to 99.28\%) on a variety of corpora such as SMS spam collections, Enron, SpamAssassin, and Ling-Spam. For SMS spam identification, \cite{liu2021spam} introduced both a customized Transformer model and a Vanilla Transformer model with modifications. They utilized the GloVe method for text message vector representations, achieving an accuracy of 98.92\%. Using a Kaggle dataset, the authors \cite{hossain2022detecting} introduced a spam detection model that used BERT to outperform both deep learning techniques and conventional machine learning algorithms, with an accuracy of 98.80\%. Notably, it optimizes hyperparameters to speed up model selection and improves model performance by training BERT with extra layers on a benchmark dataset. \cite{oswald2022spotspam} addresses dynamic keywords well by concentrating on word semantics in SMS spam filtering through the use of an intention-based approach. Different pre-trained NLP models are used to provide contextual embeddings, and 13 pre-specified intention labels are used to capture the textual and semantic components of short-text messages. By computing intention scores and using supervised learning classifiers, the model obtains impressive results on the SMS Spam Collection benchmark dataset, with an accuracy of 98.07\%. The authors \cite{tida2022universal} highlighted the vital role of self-attention mechanisms in enabling BERT models to detect spam effectively. They developed a Universal Spam Detection Model (USDM) trained on four different publicly available datasets: Enron, Spamassain, Lingspam, and Spam Text datasets. By fine-tuning with the same hyperparameters and ensuring F1-scores above 0.9 for each dataset when using the corresponding model, this approach achieved an accuracy of 97\% with an F1 score of 0.96.

The literature reviews on SMS spam detection tasks highlight existing research works that demonstrate good performance using ML, deep learning, and transformer-based models. However, although those proposed techniques and models work well, but several types of questions can arise, such as “How do the black-box models work?” and “Are the predictions of those black-box models trustworthy or not?”. To answer those questions, XAI has become a prominent field. The XAI approach improves the black-box model's explainability and transparency so that they can make more interpretable decisions. In this research work, two popular XAI techniques have been employed to answer those questions and make the proposed approach trustworthy.

\section{Methodology}

\begin{figure*}[htbp]
\centering
\includegraphics[width=0.9\linewidth]{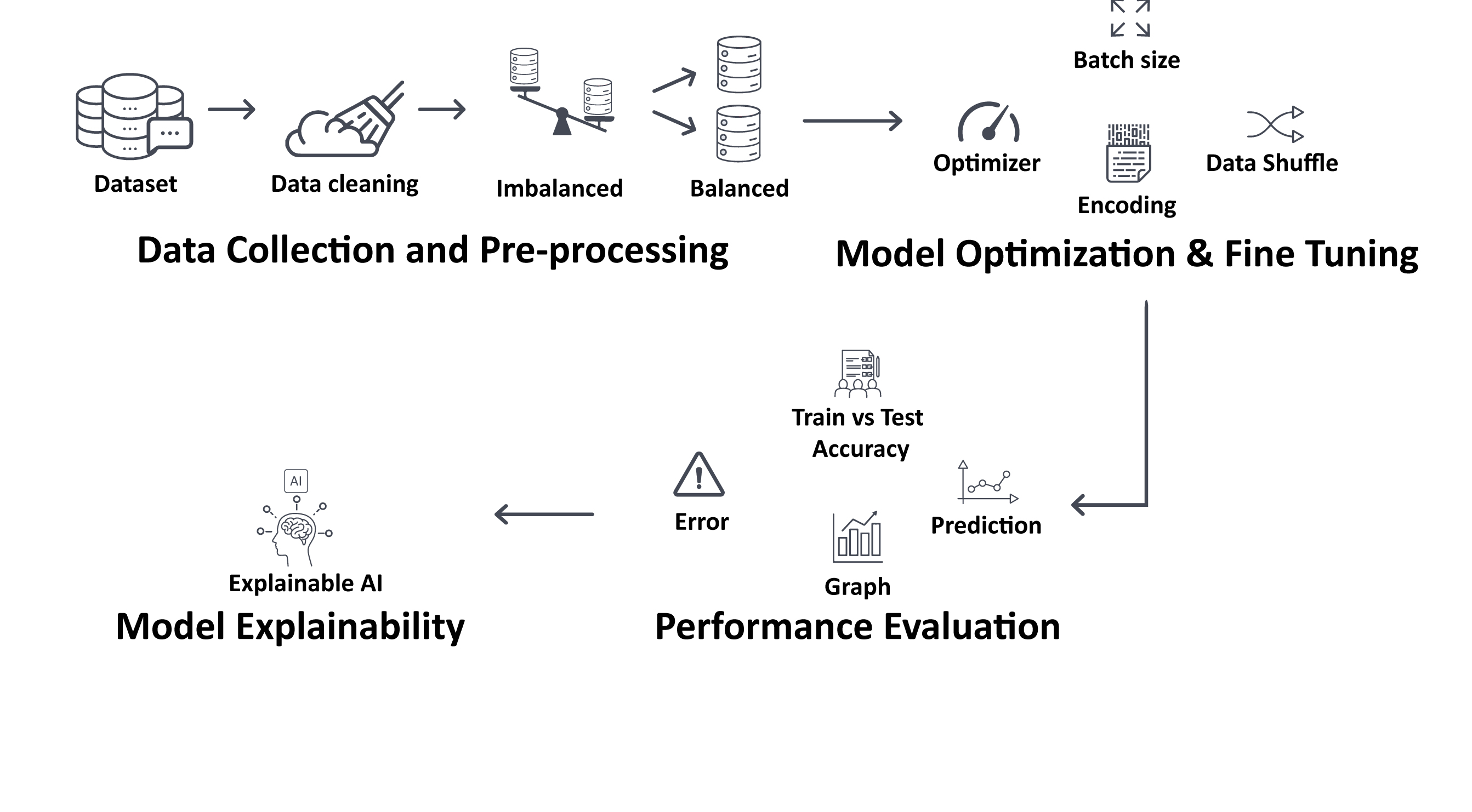}
  \caption{The methodology of SMS Spam Detection.}
  \label{fig:Methodology}
\end{figure*}

In this research work, we have tried to demonstrate the effectiveness of LLMs, such as transformer-based self-attention mechanism BERT models that can make a significant impact in the cybersecurity field, particularly in  SMS spam detection. Initially, we collected a benchmark dataset that had already been used in other research for this spam binary classification task. We prepared this dataset for further experiments using several preprocessing techniques and data balancing techniques, resulting in clean, noise-free, and balanced data. After preprocessing, for this spam detection, we selected several traditional ML models as well as two BERT variants DistilBERT and RoBERTA models. We trained those transformer-based models with the prepared dataset by applying optimization and fine-tuning mechanisms. After analyzing the performance of the ML and BERT models we showed comparisons between them.  Finally, two well-known XAI techniques LIME and Transformers Interpret are employed to explain how the fine-tuned BERT variant model is working and making decisions in this spam classification. The methodology of the research work is presented in Fig. ~\ref{fig:Methodology}.

\subsection{Data Collection}\label{subsec1}

\begin{figure*}[htbp]
    \centering
    \includegraphics[width=0.9\linewidth]{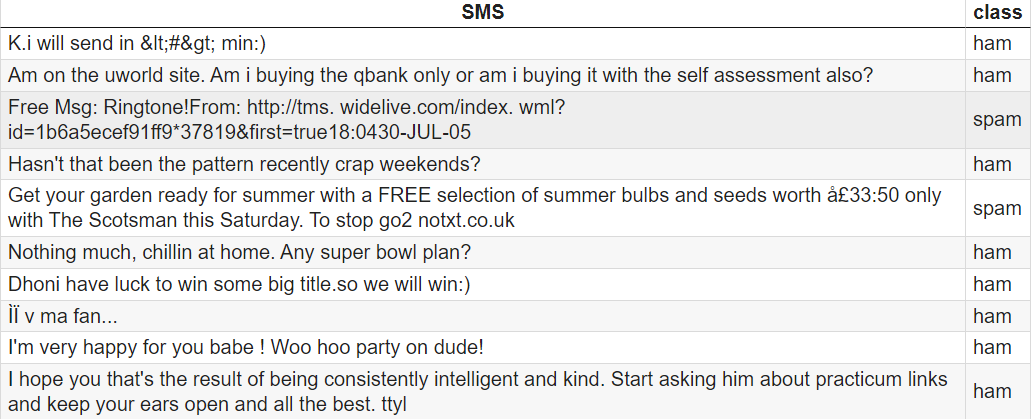}
    \caption{The sample data overview}
  \label{fig:SampleData}
\end{figure*}

A publicly available SMS spam dataset has been collected from the UCI machine learning repository for this experiment. This dataset is labeled and also collected for mobile phone spam research. There are 5,574 English SMS messages in the dataset, all of which are classified as "spam" or "ham". Fig. ~\ref{fig:SampleData} shows a sample of the data. The dataset shows that in between 5,574 data, 4825 data existed as “ham” data and 747 as “spam”. This distribution of data indicates that the dataset is not well-balanced.

\subsection{Data Preprocessing}\label{subsec2}
To prepare the dataset, we need to clean the data by solving different types of problems that exist in this data. Mainly we focused on the minor changes in the data because using several preprocessing techniques can potentially reduce the accuracy of NLP tasks. Several preprocessing techniques such as punctuation, and stopword removal can inadvertently remove important information from the text.  For this spam detection task, we solely clean the data by replacing non-alphanumeric characters in the text with a space. Afterward, we replaced all whitespace characters, such as spaces, tabs, and newlines, with a single space. Then, we removed all carriage characters, replacing them with spaces. Finally, we removed all repeated words from the text. To increase the accuracy of the model, preprocessing the SMS data was essential since it highlighted important textual characteristics and patterns that were required to recognize spam messages. Algorithm ~\ref{algo1} shows the preprocessing process.

\begin{algorithm}
\caption{Text Preprocessing}\label{algo1}
\begin{algorithmic}[1]
\Require Text to be preprocessed
\Ensure Cleaned and preprocessed text
\Function{Clean\_Text}{text}
    \State text $\Leftarrow$ \Call{Replace\_Special\_Characters}{text}
    \State text $\Leftarrow$ \Call{Remove\_Extra\_Spaces}{text}
    \State text $\Leftarrow$ \Call{Remove\_Carriage\_Returns}{text}
    \State \Return text
\EndFunction

\Function{Remove\_Repeat}{text, repeat}
    \State words $\Leftarrow$ \Call{Split\_Text}{text}
    \State result $\Leftarrow$ []
    \For{word \textbf{in} words}
        \If{\Call{Count\_Occurrences}{result, word} $<$ repeat}
            \State \Call{Append}{result, word}
        \EndIf
    \EndFor
    \State \Return \Call{Join\_Words}{result}
\EndFunction

\Function{Preprocess\_Text}{text}
    \State text $\Leftarrow$ \Call{Clean\_Text}{text}
    \State text $\Leftarrow$ \Call{Remove\_Repeat}{text, 1}
    \State \Return text
\EndFunction
\end{algorithmic}
\end{algorithm}

\begin{figure*}[htbp]
    \centering
    \begin{subfigure}{0.4\textwidth}
        \includegraphics[width=\linewidth]{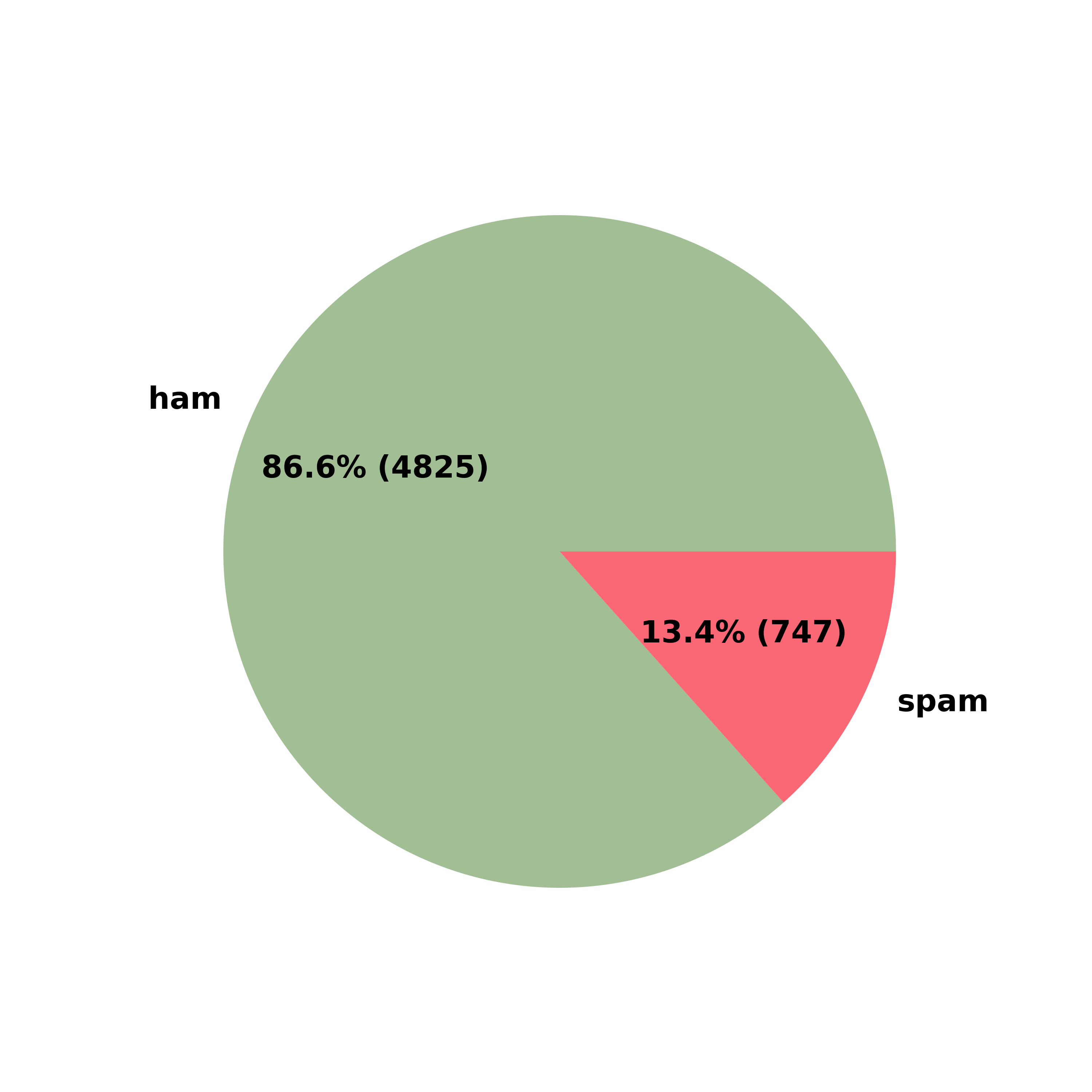}
        \caption{Imbalanced data}
    \end{subfigure}%
    \begin{subfigure}{0.4\textwidth}
        \includegraphics[width=\linewidth]{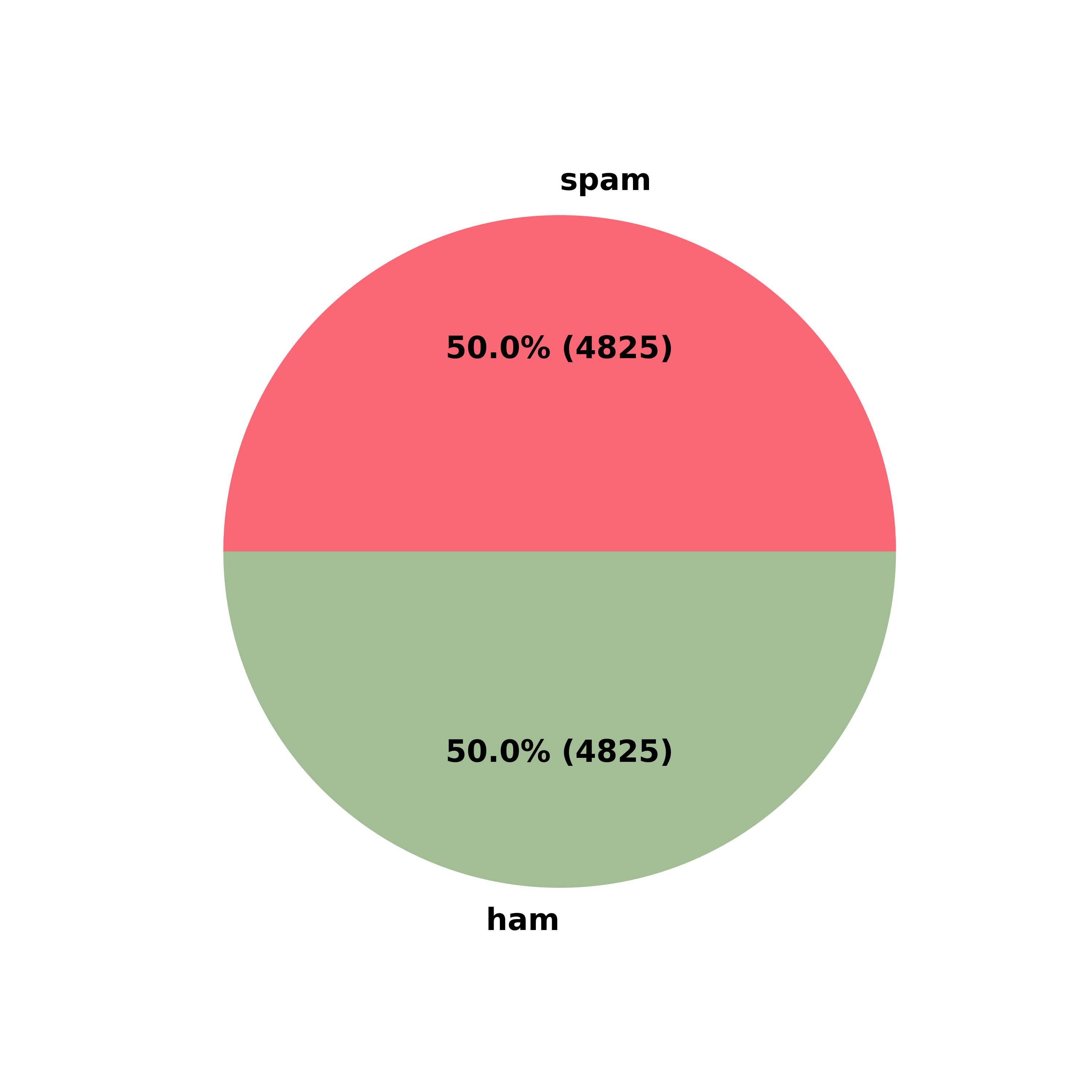}
        \caption{Balanced data}
    \end{subfigure}
    \caption{Visualizing Imbalanced vs. Balanced Data Distribution.}
    \label{fig:ImbalancedBalancedData}
\end{figure*}

After the data pre-processing phase, we employed a text augmentation technique back translation, and data resampling to balance the data. This augmentation method involves translating the original text data into one or more languages and then translating the translated text back to the original language using machine learning services like Google Translate. Back translation enhances our dataset, making it easier to train robust models, improving performance on a range of natural language processing tasks, and producing text data with different terms while maintaining context. Algorithm ~\ref{algo2} shows the process of data augmentation using back translation. As in our data “spam” is a minor class we have done overall oversampling techniques in this class. After back translation, we combined the back-translated text with the original text. Overall this oversampling process, the dataset has become balanced and the dataset has 4,825 “ham” and 4,825 “spam” data in equal portions. Unbalanced datasets can produce biased models that perform poorly in minority classes by favoring the majority class. By ensuring the model learns appropriate characteristics across all classes and improving its generalization abilities, balancing the data reduces the risk of overfitting. Fig.~\ref{fig:ImbalancedBalancedData} illustrates the distribution of imbalanced and balanced data before and after oversampling.

\begin{algorithm}
\caption{Data Augmentation using Back Translation}\label{algo2}
\begin{algorithmic}[1]
\Require Original data
\Ensure Augmented data using the back translation
\Function{Back\_Translate}{sentence, languages}
    \State translator $\Leftarrow$ \Call{Initialize\_Translator}{}
    \State target\_language $\Leftarrow$ \Call{Random\_Choice}{languages}
    \State translation $\Leftarrow$ \Call{Translate}{translator, sentence, target\_language}
    \State translated\_text $\Leftarrow$ \Call{Get\_Translated\_Text}{translation}
    \State reverse\_translation $\Leftarrow$ \Call{Translate}{translator, translated\_text, 'en'}
    \State back\_translated\_text $\Leftarrow$ \Call{Get\_Translated\_Text}{reverse\_translation}
    \State \Return back\_translated\_text
\EndFunction

\For{sentence \textbf{in} original\_data\_list}
    \State augmented\_sentence $\Leftarrow$ \Call{Back\_Translate}{sentence, [language1, language2, ...languageN]}
\EndFor
\end{algorithmic}
\end{algorithm}

\subsection{Dataset splitting}\label{subsec3}

Partitioning a dataset into several subsets is known as dataset splitting, and it is usually done to train and evaluate deep learning or machine learning models. This split is crucial to assess the model's performance on unobserved data. We may determine how effectively a model generalizes to new instances by training it on one subset and testing it on another. This helps to prevent overfitting, a phenomenon in which a model becomes adept at memorizing training data but has trouble with new data. The dataset has been divided into two parts for this experiment: the training set, which contains 80\% of the data, and the testing set, which includes the remaining 20\%. There are 7,720 data points in the training set and 1,930 data points in the testing set.

\subsection{Model Selection}\label{subsec4}

To achieve the best performance, generalization, and practicality in real-world applications, proper model selection ensures that the selected model matches the criteria, restrictions, and features of the classification or detection task. For this detection purpose, initially, we have utilized several traditional ML algorithms named Extreme Gradient Boosting (XGB) \cite{chen2016xgboost}, Support Vector Machine (SVM) \cite{wang2005support}, K-Nearest Neighbor( KNN) \cite{guo2003knn}, and Random Forest (RF) \cite{breiman2001random}. Along with the ML models, we also employed transformer-based BERT variant models such as DistilBERT and RoBERTa.The main goal of our utilizing both ML models and transformer-based models is to classify the SMS spam with high accuracy and show the best model that performed well in this detection context.

\subsubsection{DistilBERT}\label{subsubsec1}

DistilBERT is a simplified and efficient version of the BERT model. It is designed to provide comparable performance to BERT but at a much smaller, lighter, and faster pace. As a result, it is better suited for implementation in contexts with limited resources or in applications that demand faster inference. DistilBERT is still the preferred transformer-based model for many natural language processing applications because of its effectiveness. It uses a process called knowledge distillation to achieve this. It gains the ability to mimic the functionality of a larger model like BERT using a smaller dataset during training. DistilBERT significantly reduces the size and computing needs of BERT while preserving most of its performance by extracting key insights from the larger model. As a result, it reduces the model size by 40\% while retaining 97\% of BERT's language understanding ability and achieving a 60\% speed improvement \cite{sanh2019distilbert}. It uses the Transformer architecture, which has feed-forward neural networks similar to BERT and self-attention layers. However, it usually has fewer hidden sizes and fewer layers, resulting in a reduction in the number of parameters. In contrast, the BERT base model typically comprises 12 layers, while DistilBERT may have only 6 layers with a hidden size of 768 \cite{adoma2020comparative}. There are about 66 million parameters in DistilBERT, compared to 110 million in the base BERT model. DistilBERT uses tokenization, which turns raw text inputs into numerical embeddings, to handle texts. By utilizing attention mechanisms, these embeddings capture word relationships and help the model identify and prioritize important data.

\subsubsection{RoBERTa}\label{subsubsec1}

Robustly Optimized BERT Approach, or RoBERTa, is an enhanced variant of BERT, a transformer-based model that incorporates improvements to better handle larger batch sizes and train on longer sequences. Similar to BERT, RoBERTa prioritizes the encoder within the Transformer architecture for text and sentence pair classification tasks. Their approaches to tokenization are one of the main differences between the two: BERT uses the WordPiece tokenizer, whereas RoBERTa uses a byte-level Byte Pair Encoding (BPE) tokenizer, employing a larger vocabulary set comprising 50K subword units \cite{tan2022roberta}. Tokenization is also performed by RoBERTa using BPE as a component rather than character processing directly. RoBERTa learns unsupervised from a large corpus of text data during its pre-training phase.  Similar to BERT's Masked Language Model aim, it trains on predicting masked words within sentences. Additionally, to enhance the caliber of learned representations, it does away with the NSP (Next Sentence Prediction) task and employs a larger training corpus with dynamic masking. RoBERTa performs better than BERT, in part, because of this change in training objectives. Its improved performance is also a result of training methodology optimizations, which include bigger batch sizes, longer training times, and dynamic masking.

\subsection{Model Optimization and Fine Tuning}\label{subsec5}

Model optimization aims to reduce complexity and increase efficiency, whereas fine-tuning modifies current models to fit new datasets or tasks, improving performance via transfer learning. Both procedures are essential for improving transformer-based models so they perform well in a variety of scenarios. The Transformers tokenizer is used to tokenize the preprocessed spam dataset. It segments text into smaller units using a sub-word-based technique called WordPiece or Byte Pair Encoding, which helps the model better understand word meaning and context. In this tokenization procedure, we initialized the maximum sequence length of the text to 512. Tokenization relies on two key approaches: padding and truncation are essential techniques used to ensure that input sequences have a consistent length. This is particularly important when working with neural networks, as they typically require fixed-size inputs. Padding involves adding tokens (typically zeros) to sequences that are shorter than the maximum length, whereas truncation involves removing tokens from longer sequences. 

The choice of optimal batch size is crucial in deep learning, impacting both training and testing accuracies as well as overall runtime \cite{radiuk2017impact, lin2022analysis, schmeiser1982batch}. This selection involves a trade-off between training speed and memory usage. In this study, a batch size is selected to be 32 for training data, while a batch size of 64 is used for testing data. In order to avoid overfitting and ensure exposure to a wide range of previously unseen samples, the training data is shuffled at each epoch. The model's capacity for generalization is increased and its ability to generalize is strengthened by this randomization. Additionally, pre-trained weights from the pre-training phase initialize the DistilBERT and RoBERTa models, optimizing performance.

An optimizer is an essential part of neural networks that updates pre-trained model weights. An optimizer's primary objective is to minimize a loss function by changing the model's parameters during training. This experiment uses AdamW (Adam Weight Decay) \cite{loshchilov2017decoupled}, an effective and popular optimization approach, to update the pre-trained model's weights. The AdamW optimizer is a modified and improved version of Adam that is widely used for deep neural network training. It decouples the regularizer's gradient from the Adam-$\ell_2$ update rule, making it effective at minimizing a loss function and lowering overfitting \cite{zhuang2022understanding}. Weight decay, often called L2 regularization, is a regularization technique that is commonly used in neural network training to reduce overfitting. It entails adding a penalty term to the loss function to prevent the use of too high parameter values.

The learning rate is an essential hyperparameter that significantly impacts the dynamics of training and model performance. It determines the size of the steps are in the optimization process, which affects convergence and overall efficacy. The ideal learning rate selection is contingent upon several elements, such as the specific task at hand, model design, and optimization techniques. A high learning rate might cause instability, which makes it difficult to generalize to new data. Conversely, a low learning rate could make training take longer, costing more computational resources since more epochs would be needed to reach the target performance levels. For training the DistilBERT and RoBERTa models in our work, we employ a learning rate of 2e-5 (0.00002), which is a typically used value for transformer-based architectures such as BERT and its derivatives. By balancing the trade-off between model performance and training speed, this decision ensures effective convergence while lowering the possibility of instability. For model optimization and fine-tuning the hyperparameter settings for the suggested transfomer-based models are shown in Table \ref{tbl1}.

\begin{table}[!htbp]
\caption{Hyper-parameter Settings for SMS spam Detection.}\label{tbl1}
\begin{tabularx}{\linewidth}{@{} llc @{}}
\toprule
Parameter & Utilized Values & Optimal Value\\
\midrule
Training Batch size & 8, 16, 32, 64 & 32 \\
Testing Batch size & 8, 16, 32, 64 & 64 \\
Optimizer & AdamW & AdamW \\
Learning rate & 1e-5, 2e-5, 3e-5, 4e-5 & 2e-5 \\
Number of epochs & 2, 3, 4, 5, 6, 7 & 5 \\
\bottomrule
\end{tabularx}
\end{table}

\begin{figure}[htbp]
    \centering
    \includegraphics[width=\linewidth]{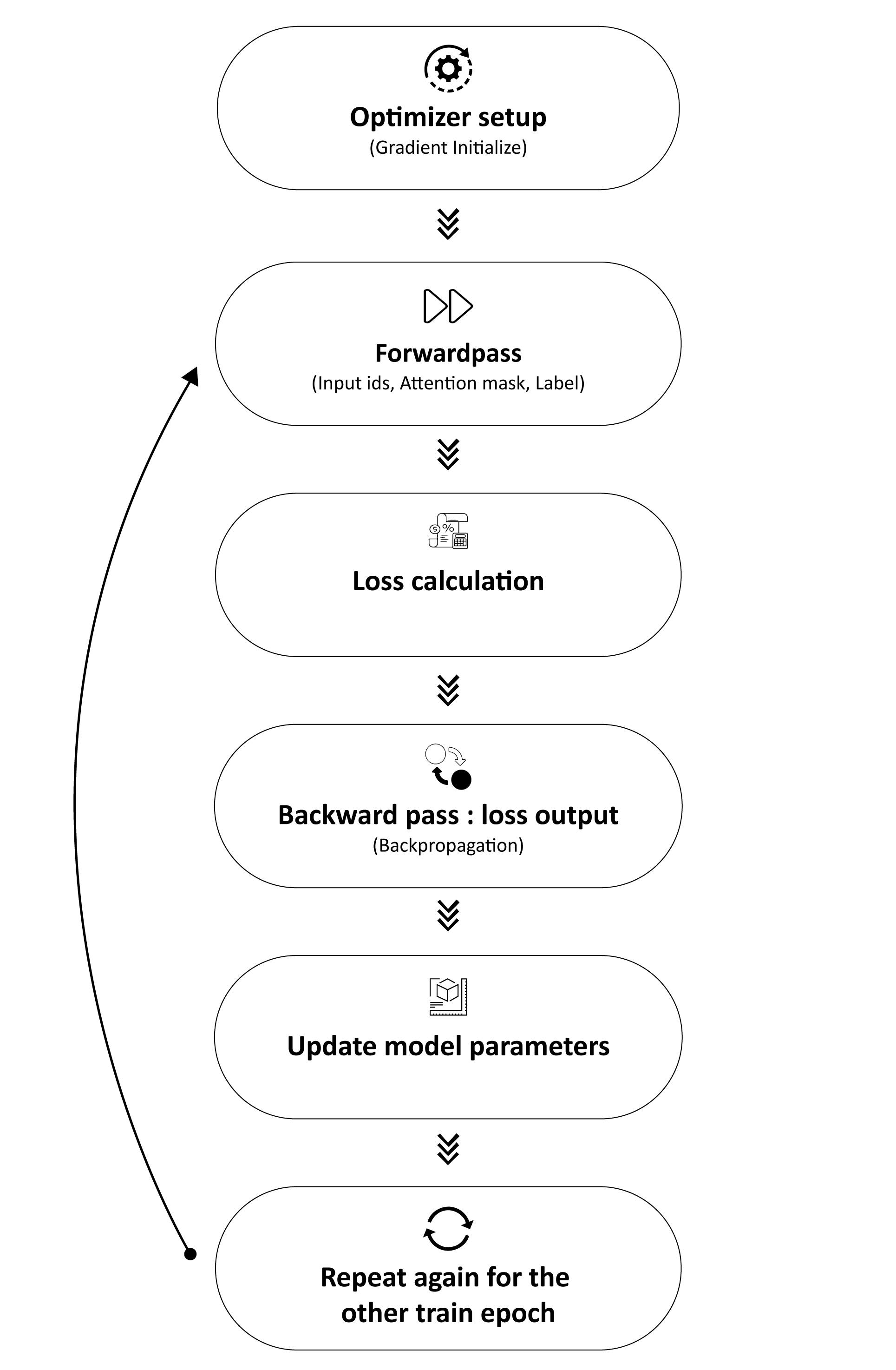}
    \caption{The Complete Fine-Tuning Process Flow.}
    \label{FIG:Fine_Tuning}
\end{figure}

After optimizing the model through hyperparameter tuning, the fine-tuning process is implemented to train the model on specific tasks and datasets. This phase involves critical hyperparameters including batch size, regularization methods, and learning rate. In fine-tuning, an optimization algorithm and loss function are selected, and backpropagation is used to iteratively alter the parameters of the entire model or selected layers.  The optimization process modifies the parameters to reduce the difference between the model's predictions and the dataset's ground truth labels. In this experiment, 80\% of the entire data was used for fine-tuning. The training data is sent through the models in batches throughout each epoch, and gradients are calculated via the backpropagation technique. To modify pre-trained models for the given task such as SMS spam detection, RobertaForSequenceClassification and DistilBERTForSequenceClassification were utilized. The train\_loader and test\_loader were configured using DataLoader. The train\_loader was configured by preparing the data, figuring out the batch size, and reorganizing the data among epochs. However, no shuffling was involved in configuring the test loader. The overall fine-tuning process flow diagram, featuring these procedures for DistilBERT and RoBERTa, is displayed in Fig. \ref{FIG:Fine_Tuning}. 

After setting up the optimizer with its parameters and learning rate, the model receives input parameters during the forward pass to generate predictions. Then, using automatic differentiation, the gradients of the loss with respect to the model's parameters such as weights and biases are calculated in the backward pass. This determines the loss function's gradients in relation to each parameter. The optimizer uses these gradients to update the model's parameters during the optimization process. This overall process iterates until all the epochs of the model training conclude. As a result, the model can recognize important patterns and features in the training set. Once the training is complete, the fine-tuned model is ready to predict the spam messages with unseen new data. Algorithm \ref{algo3} shows the fine-tuning steps of the optimized transformer-based model with the train data.

\begin{algorithm}
\caption{Fine-Tuning Transformer-based (DistilBERT, RoBERTa) Model}\label{algo3}
\begin{algorithmic}[1]
\Require Pre-trained model, train\_loader, optimizer, learning rate, maximum training epochs
\Ensure Fine-tuned Trained Transformer-based model

\State $\text{epochs} \gets \text{max\_epochs}$
\State $\text{optimizer} \gets \text{Optimizer}(\text{model.parameters()}, \text{learning\_rate})$
\State $\text{device} \gets \text{cuda}$ 

\Function {FineTuneTransformerModel}{model, train\_loader, optimizer, epochs}
\For{$\text{epoch}$ in $\text{range(epochs)}$}
    \State $\text{model.train()}$
    \For{$\text{batch}$ in $\text{train\_loader}$} 
        \State $\text{batch\_input\_ids} \gets \text{batch}[0].\text{to(device)}$ 
        \State $\text{batch\_attention\_mask} \gets \text{batch}[1].\text{to(device)}$
        \State $\text{batch\_labels} \gets \text{batch}[2].\text{to(device)}$
        \State $\text{optimizer.zero\_grad()}$ \Comment{Clear gradients}
        \State $\text{outputs} \gets \text{model}(\text{input\_ids}\gets\text{batch\_input\_ids}, \text{attention\_mask}\gets\text{batch\_attention\_mask}, \text{labels}\gets\text{batch\_labels})$ \Comment{Forward pass}
        \State $\text{calculated\_loss} \gets \text{outputs.loss}$ 
        \State $\text{calculated\_loss.backward()}$ \Comment{Backward pass}
        \State $\text{optimizer.step()}$ \Comment{Update parameters}
    \EndFor
\EndFor
\State \textbf{return} $\text{model}$
\EndFunction
\end{algorithmic}
\end{algorithm}

\subsection{Model Explainability}\label{subsec6}

\begin{figure}[htbp]
\centering
\includegraphics[width=\linewidth]{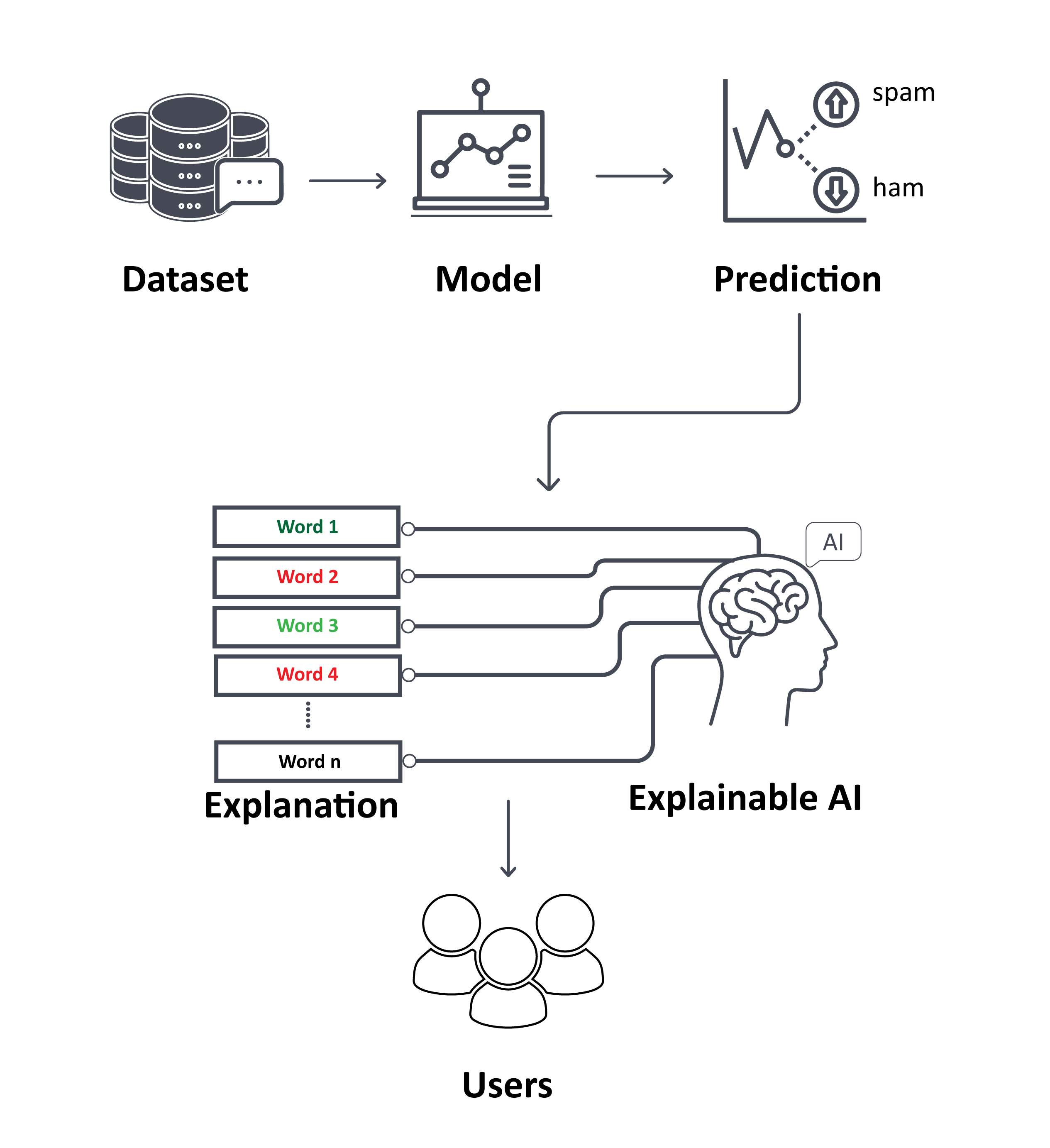}
  \caption{The complete approach to understanding model explainability}
  \label{fig:Model_Explainability}
\end{figure}

Explainability of a model refers to the ability to comprehend and analyze the process by which a model makes its predictions or decisions. A model can be very accurate across various tasks, but it often behaves like a black box, making it challenging to understand why it predicts specific outcomes. Approaches for elucidating black-box machine learning models aim to enhance transparency, yet their explanations may harbor significant ambiguity, leading to diminished user confidence and raising concerns regarding the model's fairness and reliability \cite{zhang2019should}. XAI aims to develop machine learning methods that enable people to understand and trust the judgments made by complex models, especially deep neural networks, which frequently operate as opaque black boxes \cite{holzinger2022explainable}. The necessity for greater openness in these models may impede the adoption and trustworthiness of AI and ML technologies. This issue is addressed by XAI, which offers techniques for interpreting and explaining model predictions. An in-depth understanding of XAI's fundamental concepts and familiarity with related programming frameworks are essential for selecting the appropriate methodology when developing an application that leverages XAI \cite{dwivedi2023explainable}. The XAI community has developed numerous effective methods for explaining predictions, which is crucial for fostering acceptance and confidence in AI and ML technologies. In this research, we utilized two Explainable Models, LIME and Transformers Interpret, to understand the model predictions and address the black box problem. The comprehensive process of model explainability is illustrated in Fig. \ref{fig:Model_Explainability}.

The experiment for SMS spam detection is a text classification challenge in which LimeTextExplainer is used to explain how the black box transformer-based model operates.  It is an interpretable explanation approach that builds a local interpretable model around a prediction to make any classifier's predictions more understandable \cite{ribeiro2016should}. To achieve model explainability, LIME initiates the process by perturbing the SMS content through subtle modifications such as rearranging or removing words or adding new ones. This perturbation investigates how sensitive the model is to changes in the input data. Then, the modified SMS content is fed into the transformer-based black-box model to produce predictions. LIME extracts prominent characteristics from the resulting instances that are affected by the perturbations. These local models simulate how the black-box model would behave close to the chosen instance and offer insights into how it makes decisions. After that, LIME uses the knowledge gathered from the local model to explain the predictions made by the black-box model on the original dataset. These justifications usually highlight the terms or critical components that substantially impact the model's decision-making process, improving the predictive model's overall interpretability. By taking a thorough and iterative approach, LIME helps to clarify the complicated inner workings of machine learning models, increasing transparency and confidence in their forecasts. Eq. \ref{eq:lime} shows the formulation used by LIME to generate an interpretable explanation for a specific data point \cite{molnar2020interpretable}.

 \begin{equation}\label{eq:lime}
\begin{aligned}
\text{LIME\_exp}(n) &= \text{argmin}_{x \in \text{X}} \left( L(m, x, \pi_n) + \Omega(x) \right)
\end{aligned}
\end{equation}

In this context, m denotes the black box model that predicts results based on input data n. The LIME model aims to generate an understandable explanation for a particular data point n, denoted as x. An interpretable model x approximates the behavior of the black-box model m locally around n, resulting in the explanation. The hypothesis space of a specific interpretable learner is represented by X. L(m, x, $\pi_n$) measures the average loss between the predictions of the interpretable model x and the black-box model m for data points in the neighborhood of $\pi_n$. In the context of local approximation, $\pi_n$ represents the perturbations or features around n that are considered. These perturbations aid in comprehending the behavior of the black-box model near n. $\Omega$(x) represents a regularization term or penalty applied to the complexity of the interpretable model x. This term penalizes overly complicated models, encouraging simpler explanations.

In order to clarify the operation of the transformer-based model, a Transformers Interpret tool was also included in the categorization process. This tool uses a class explainer technique that uses a tokenizer and pre-trained model to help explain model predictions. Word attributions are computed as part of the class explainer process to reveal how individual words contribute to the model's predictions. The terms that significantly influence the model's decision-making are identified using word attribution analysis. Positive scores represent a word's influence that is positive on the predicted class, and negative scores represent a negative term. In addition, this procedure's explanation highlights terms that significantly impact the model's decision-making. The words that have been highlighted function as crucial identifiers of the attributes that the model gives priority to during the prediction process. The procedure for creating explanations for model predictions using this class explainer approach is described in detail in Algorithm \ref{algo4}. By using this technique, users can enhance transparency and trust in their predictions, gaining a deep understanding of the inner workings of transformer-based models.

\begin{algorithm}
\caption{Explainability Analysis using XAI}\label{algo4}
\begin{algorithmic}[1]
\Require model, tokenizer, input text
\Ensure Explanation with Word attributions for model predictions
\State class\_names $\Leftarrow$ [Class1, Class2, Class3 ..., ClassN]

\Function{Explain\_Model\_Prediction}{model, tokenizer, input\_text}
    \State $lime\_exp \Leftarrow$ \Call{ExplainWithLIME}{model, input\_text, num\_features, top\_labels}
    \State $transformer\_exp \Leftarrow$ \Call{Transformer\_Interpretation}{model, tokenizer, input\_text}
    \State \Return $lime\_exp$, $transformer\_exp$
\EndFunction

\Function {ExplainWithLIME}{model, input\_text, num\_features, top\_labels}
    \State $lime\_explainer \Leftarrow$ LimeTextExplainer(class\_names)
    \State $lime\_explanation \Leftarrow$ $lime\_explainer$.explain\_instance (input\_text, model, num\_features, top\_labels)
    \State $label \Leftarrow$ lime\_explanation.get\_label()
    \State $explanation\_attributions \Leftarrow$ explanation.as\_list(label)
    \State $\text{visualize\_explanation}()$
    \State \Return $explanation\_attributions$
\EndFunction

\Function{Transformer\_Interpretation}{model, tokenizer, input\_text}
    \State $trans\_explainer$ $\Leftarrow$ SequenceClassificationExplainer(model, tokenizer)
    \State attributions $\Leftarrow$ $trans\_explainer$(input\_text)
    \State $\text{visualize\_interpretation}()$
    \State \Return $attributions$
\EndFunction
\end{algorithmic}
\end{algorithm}

\section{Result Analysis}

In this section, We describe in detail the proposed method's comprehensive outcomes and provide a proper comparison between traditional ML models (XGB, SVM, KNN, and RF) utilized in this work. The experiment uses several key metrics such as precision, recall, F1-score, and accuracy to analyze the results of a classification model. Our utilized dataset is significantly imbalanced because the number of spam messages is less than that of ham messages. In this experiment, we tested our model with an unbalanced and balanced dataset to see the model’s effectiveness. To evaluate the performance of the fine-tuned RoBERTa model on imbalanced and balanced data those key metrics have been calculated.
To summarize the overall performance of a model, calculating the confusion matrix is essential as it forms the foundation for all subsequent key metrics that are derived directly from it. It is a 2x2 table or performance measuring tool that overviews how effectively a classification model has made predictions. This confusion matrix consists of four entries True Positive (TP), True Negative (TN), False Positive (FP), and False Negative (FN). A true positive (TP) in a model's evaluation happens when the model accurately predicts a positive outcome and the actual class is positive, proving that positive examples can be accurately identified. Cases where the model accurately predicts a negative outcome and the actual class is negative are referred to be true negative (TN), demonstrating accurate identification of negative events. When the model predicts a positive outcome when the actual class is negative, this is known as a false alarm, or false positive (FP). Lastly, a false negative (FN) occurs when a positive class is observed despite the model predicting a negative outcome.  These components of the confusion matrix are the main elements that are used to calculate the Precision, Recall, and F1-score that help evaluate the model's performance. 

Precision: Precision, also known as positive predictive value, is the ratio of accurately anticipated positive instances to the total number of predicted positive instances. It measures the model's accuracy in predicting positive outcomes and indicates the likelihood that a predicted positive occurrence is indeed positive. The precision is defined as follows:
\begin{equation}
\begin{aligned}
\text{Precision (P)} &= \frac{TP}{TP + FP}
\end{aligned}
\end{equation}
Recall: Recall, also known as sensitivity, quantifies the proportion of accurately anticipated positive cases to the total actual positive cases in the dataset. It measures the model's ability to detect positive samples and estimates the percentage of real positive cases that the model correctly identifies. The recall is defined as follows:
\begin{equation}
\begin{aligned}
\text{Recall (R)} &= \frac{TP}{TP + FN}
\end{aligned}
\end{equation}
F1-score: The F1-score is a well-rounded performance metric since it uses the harmonic mean to balance recall and precision. A low precision or recall will result in a lower F1-score. It offers a comprehensive assessment of a model's ability to accurately classify positive events by considering the risks of both false positives and false negatives. The F1-Score is defined as follows:
\begin{equation}
\begin{aligned}
\text{F1-score} &= 2 \times \frac{\text{P} \times \text{R}}{\text{P} + \text{R}}
\end{aligned}
\end{equation}

Accuracy: Accuracy measures how frequently the model's predictions agree with the actual labels in the dataset. It is computed by dividing the sum of true positives and true negatives by the total number of samples in the dataset. Although accuracy sheds light on the model's overall performance, an uneven class distribution may cause it to underrepresent performance. The accuracy is defined as follows:
\begin{equation}
\text{Accuracy} = \frac{\text{TN} + \text{TP}}{\text{TN} + \text{TP} + \text{FN} + \text{FP}}
\end{equation}

\subsection{Result Analysis using Imbalanced Dataset}\label{subsec1}

\begin{figure*}[htbp]
    \centering
    \includegraphics[width=0.7\textwidth]{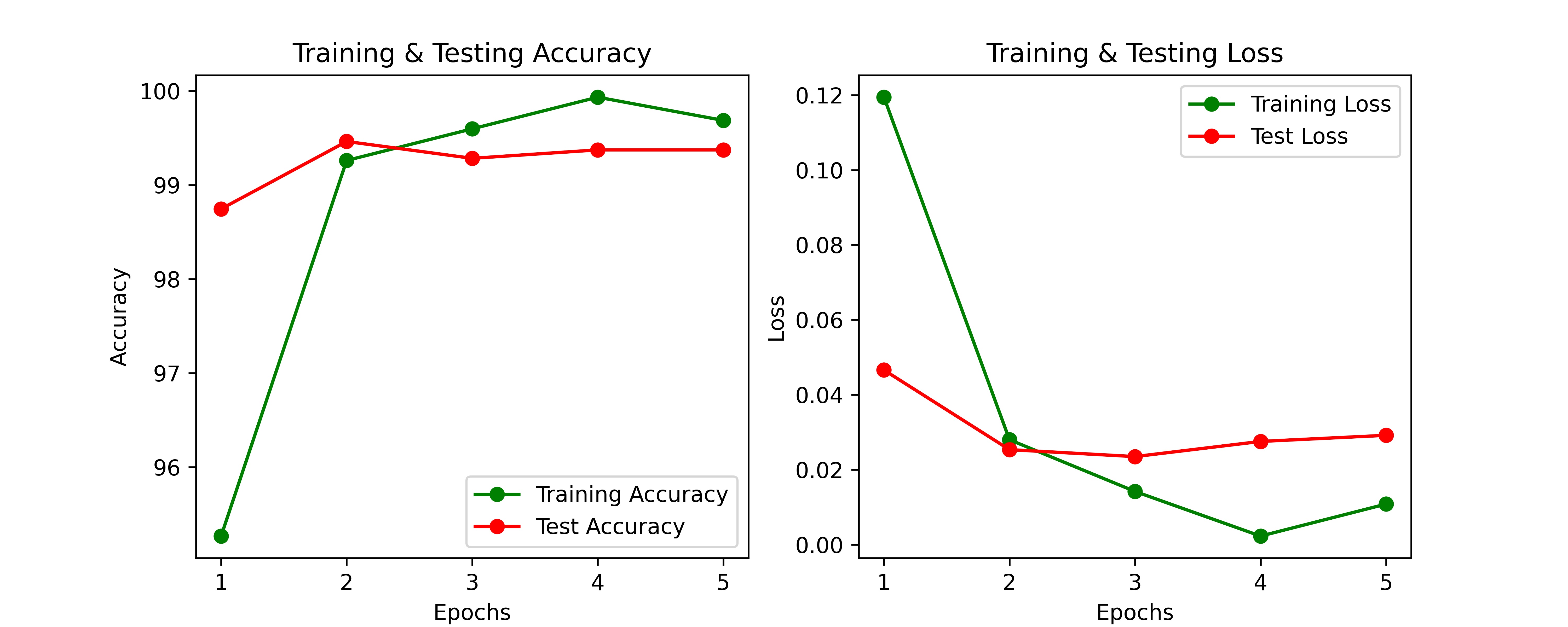}
    \caption{Training \& Testing Accuracy and Loss with Imbalanced Data}
  \label{fig:Imbalanced_loss_and_accuracy}
\end{figure*}

\begin{figure}[htbp]
    \centering
    \includegraphics[width=\linewidth]{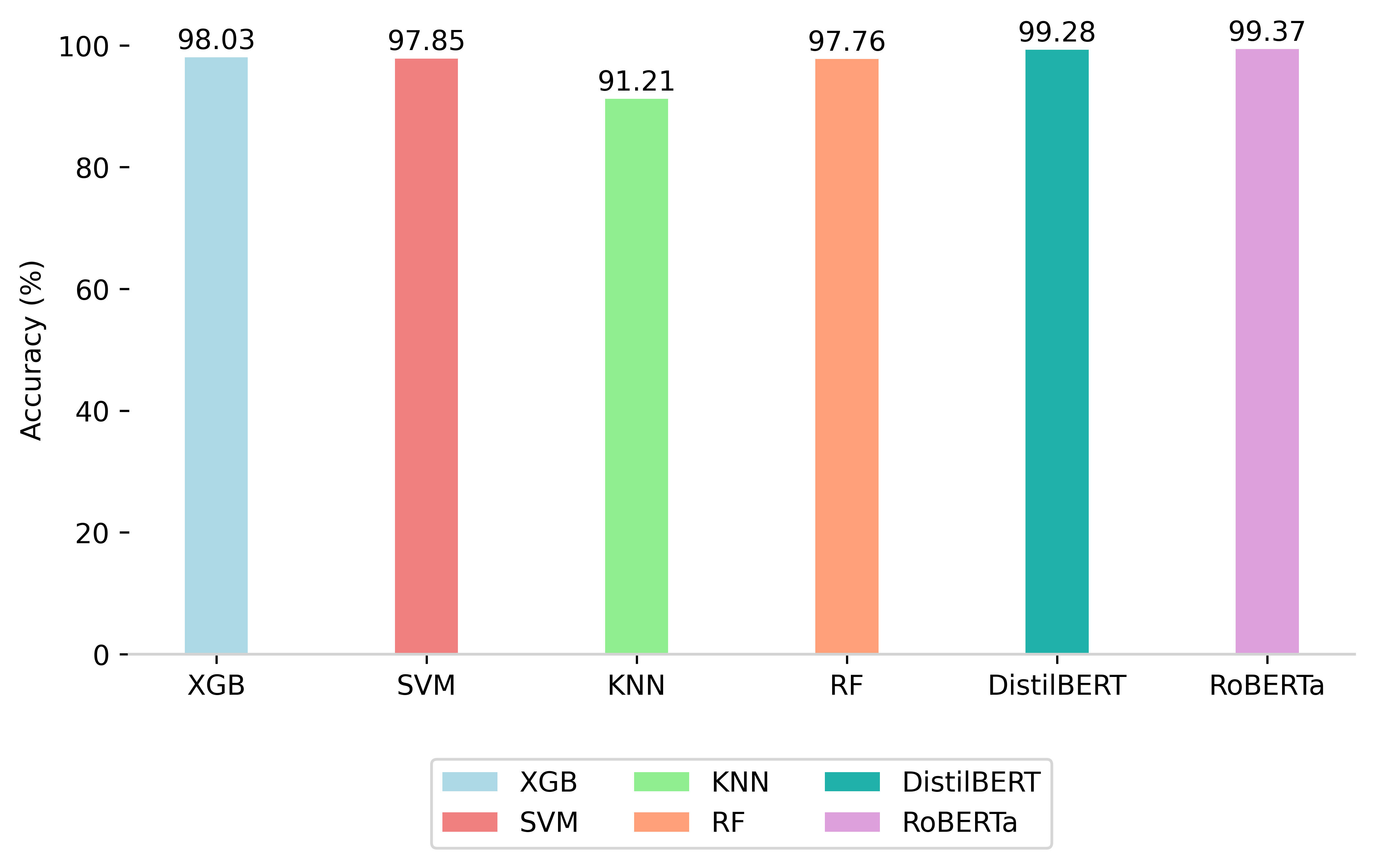}
    \caption{Representation of graphical comparison among models on imbalanced dataset}
  \label{fig:acc-imbalanced}
\end{figure}

The performance of a model can vary based on different parameters. Initially, this study used an imbalanced dataset to evaluate the model's performance during training and testing. The dataset was significantly imbalanced, with a much higher number of ham SMS than spam SMS. Classifiers such as traditional machine learning and transformer-based models were used to classify the text messages into spam and ham classes. The models were trained on the imbalanced dataset using 80\% (4,457) of the data and tested on 20\% (1,115) of the data. The results obtained from the ML and transformer-based models were impressive for the ham class; however, they were less successful for the spam class due to a lack of data. Table \ref{tbl2} displays the confusion matrix of the traditional ML and transformer-based models for the imbalanced dataset. Comparatively, among the ML models, XGB performed better than SVM, KNN, and RF. While XGB performed well, the RoBERTa model outperformed the ML and transformer-based models, providing the best overall performance. The ML and transformer models' F1-score, recall, and precision were evaluated for the imbalanced dataset, as shown in Table \ref{tbl3}. The RoBERTa classifier produced precision, recall, and F1-score values of 0.99, 1.00, and 1.00 for the ham class.
On the other hand, the model obtained precision, recall, and F1-score values for the spam class of 0.99, 0.97, and 0.98, respectively. To achieve good accuracy, we trained our optimized and fine-tuned RoBERTa model using 5 epochs with a batch size of 32. After 5 epochs, the model achieved a training accuracy of 99.69\% with a training loss of 0.0108. For the test set, it achieved a testing accuracy of 99.37\% with a testing loss of 0.0292, using a batch size of 64. Fig. \ref{fig:Imbalanced_loss_and_accuracy} presents a graph showing the training and testing accuracy and loss over 5 epochs for the imbalanced dataset. Fig. \ref{fig:acc-imbalanced}  and Table \ref{tbl4} illustrate the performance comparison among the algorithms.

\begin{table*}
\caption{Confusion matrix of ML and transformer-based models for imbalanced dataset}\label{tbl2}
\begin{tabular*}{\textwidth}{@{\extracolsep{\fill}}lcccccccccccc}
\toprule
& \multicolumn{8}{c}{ML Models} & \multicolumn{4}{c}{Transformer-based Models} \cr
\cmidrule(lr){2-9} \cmidrule(lr){10-13}
& \multicolumn{2}{c}{XGB} & \multicolumn{2}{c}{SVM} & \multicolumn{2}{c}{KNN} & \multicolumn{2}{c}{RF} & \multicolumn{2}{c}{DistilBERT} & \multicolumn{2}{c}{RoBERTa} \cr
\cmidrule(lr){2-3} \cmidrule(lr){4-5} \cmidrule(lr){6-7} \cmidrule(lr){8-9} \cmidrule(lr){10-11} \cmidrule(lr){12-13}
Class Name & ham & spam & ham & spam & ham & spam & ham & spam & ham & spam & ham & spam \cr
\midrule
ham & 963 & 2 & 963 & 2 & 665 & 0 & 965 & 0 & 961 & 4 & 963 & 2 \cr
spam & 20 & 130 & 22 & 128 & 98 & 52 & 25 & 125 & 4 & 146 & 5 & 145 \cr
\bottomrule
\end{tabular*}
\end{table*}

\begin{table}
\caption{Precision, Recall, and F1-Score of ML and Transformer-based Models for Imbalanced Dataset}
\label{tbl3}
\begin{tabular}{lcccc}
\toprule
Model & Class & Precision & Recall & F1-score \\
\midrule
\multirow{2}{*}{XGB} & ham & 0.98 & 1.00 & 0.99 \\
                     & spam & 0.98 & 0.87 & 0.92 \\
\midrule
\multirow{2}{*}{SVM} & ham & 0.98 & 1.00 & 0.99 \\
                     & spam & 0.98 & 0.85 & 0.91 \\
\midrule
\multirow{2}{*}{KNN} & ham & 0.91 & 1.00 & 0.95 \\
                     & spam & 1.00 & 0.35 & 0.51 \\
\midrule
\multirow{2}{*}{RF} & ham & 0.97 & 1.00 & 0.99 \\
                     & spam & 1.00 & 0.83 & 0.91 \\
\midrule
\multirow{2}{*}{DistilBERT} & ham & 1.00 & 1.00 & 1.00 \\
                            & spam & 0.97 & 0.97 & 0.97 \\
\midrule
\multirow{2}{*}{RoBERTa} & ham & 0.99 & 1.00 & 1.00 \\
                         & spam & 0.99 & 0.97 & 0.98 \\
\bottomrule
\end{tabular}
\end{table}

\begin{table}[width=\linewidth,cols=5]
\caption{Comparative Performance Analysis of Fine-tuned RoBERTa Model with Other Models on Imbalanced Data}\label{tbl4}
\begin{tabular*}{\tblwidth}{@{} p{2cm}p{1cm}p{1.5cm}p{1cm}p{1.5cm}@{} }
\toprule
Model & Training Loss  & Training Accuracy & Testing Loss & Testing Accuracy\\
\midrule
XGB& 0.0095 & 99.84 & 0.0810 & 98.03\\
\midrule
SVM& 0.0077 & 99.62 & 0.0591 & 97.85\\
\midrule
KNN& 0.1193 & 91.88 & 0.5429 & 91.21\\
\midrule
RF& 0.0199 & 100 & 0.1317 & 97.76\\
\midrule
DistilBERT& 0.0034 & 99.96 & 0.0343 & 99.28\\
\midrule
RoBERTa& 0.0108 & 99.69 & 0.0292 & 99.37\\
\bottomrule
\end{tabular*}
\end{table}

\subsection{Result Analysis using Balanced Dataset}\label{subsec2}
\begin{figure*}[htbp]
\centering
\includegraphics[width=0.7\textwidth]{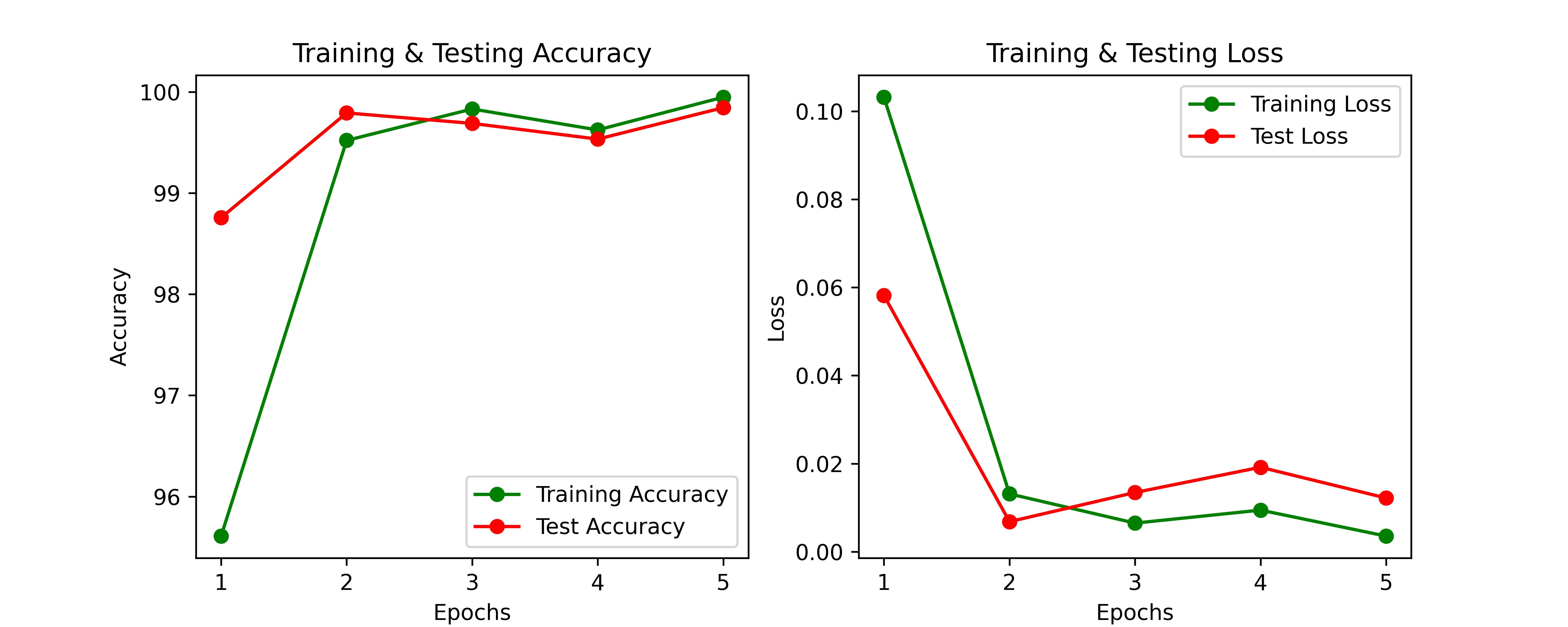}
  \caption{Training \& Testing Accuracy and Loss on Balanced Data.}
  \label{fig:Balanced_loss_and_accuracy}
\end{figure*}

\begin{figure}[htbp]
    \centering
    \includegraphics[width=\linewidth]{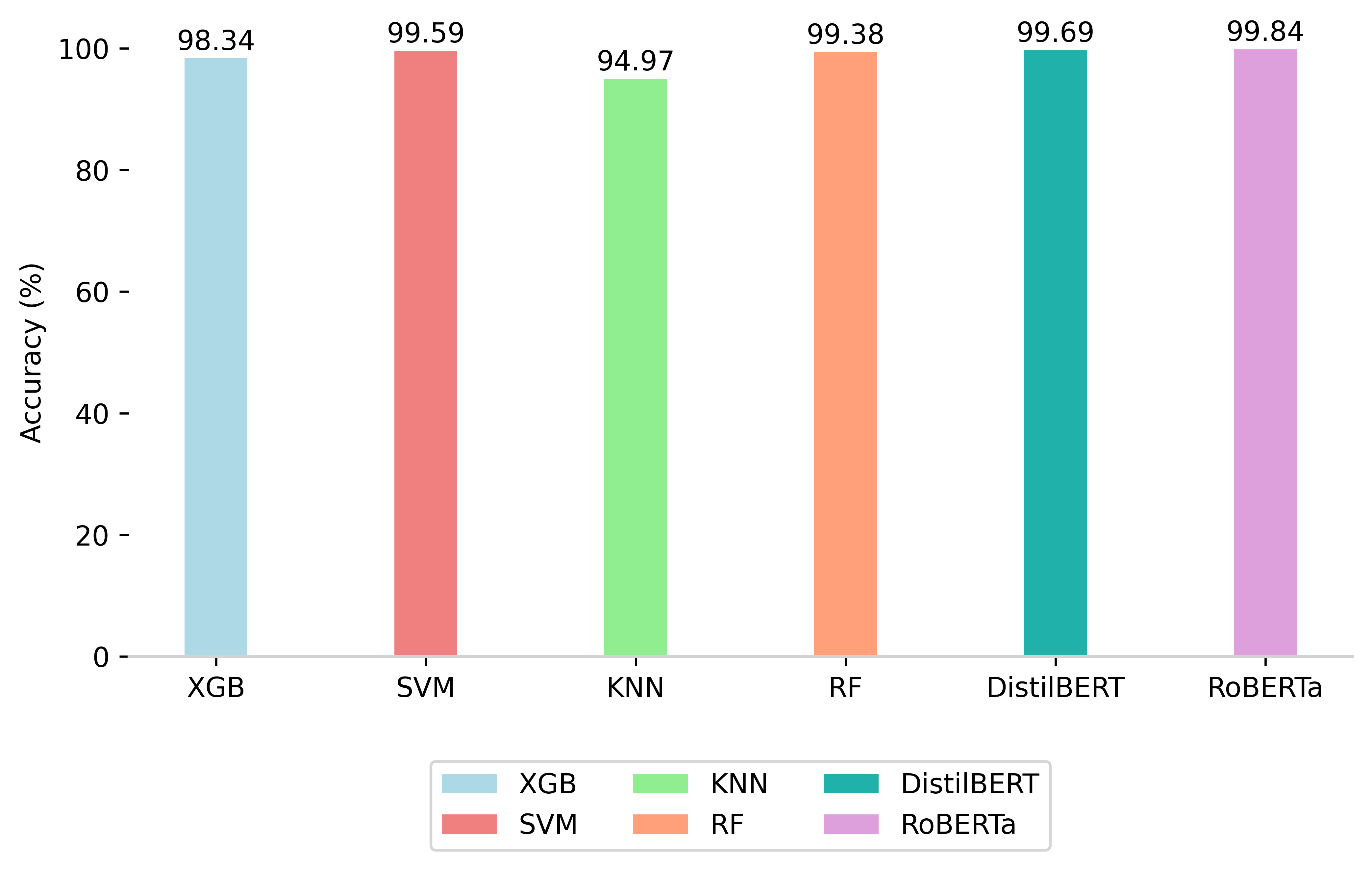}
    \caption{Representation of graphical comparison among models on balanced dataset}
  \label{fig:acc-balanced}
\end{figure}

We also experimented using a balanced dataset. Similar to the imbalanced dataset, we used 80\% (7,720 samples) of the data for training and 20\% (1,930 samples) for testing. The results from the ML and transformer-based models were impressive for both ham and spam classes. Table \ref{tbl5} shows the confusion matrix of the traditional ML and transformer-based models for the balanced dataset. In this balanced dataset analysis, the SVM algorithm performed notably well among the ML algorithms. Similarly, again our fine-tuned RoBERTa model excelled among the ML and transformer-based models. Table \ref{tbl6} presents the F1-score, recall, and precision evaluation results for the ML and transformer-based models on the balanced dataset. For the ham class, the model achieved precision, recall, and F1-score values of 1.00, 1.00, and 1.00, respectively. For the spam class, the RoBERTa classifier generated F1-score, recall, and precision values of 1.00, 1.00, and 1.00, respectively. With the balanced data, the fine-tuned model achieved 99.95\% training accuracy with a training loss of 0.0035 at the 5th epoch. Subsequently, testing the trained model resulted in a testing accuracy of 99.84\% and a testing loss of 0.0122. This demonstrates that the model's performance with a balanced dataset is significantly better than with an imbalanced dataset. Fig. \ref{fig:Balanced_loss_and_accuracy} displays a graph showing the balanced dataset's accuracy and loss over 5 epochs. Fig. \ref{fig:acc-balanced} and Table \ref{tbl7} illustrate the performance comparison among the algorithms.

\begin{table*}[htbp]
\caption{Confusion matrix of ML and transformer-based models for balanced dataset}\label{tbl5}
\begin{tabular*}{\textwidth}{@{\extracolsep{\fill}}lcccccccccccc}
\toprule
& \multicolumn{8}{c}{ML Models} & \multicolumn{4}{c}{Transformer-based Models} \cr
\cmidrule(lr){2-9} \cmidrule(lr){10-13}
& \multicolumn{2}{c}{XGB} & \multicolumn{2}{c}{SVM} & \multicolumn{2}{c}{KNN} & \multicolumn{2}{c}{RF} & \multicolumn{2}{c}{DistilBERT} & \multicolumn{2}{c}{RoBERTa} \cr
\cmidrule(lr){2-3} \cmidrule(lr){4-5} \cmidrule(lr){6-7} \cmidrule(lr){8-9} \cmidrule(lr){10-11} \cmidrule(lr){12-13}
Class Name & ham & spam & ham & spam & ham & spam & ham & spam & ham & spam & ham & spam \cr
\midrule
ham & 932 & 13 & 941 & 4 & 944 & 1 & 943 & 2 & 943 & 2 & 945 & 0 \cr
spam & 19 & 966 & 4 & 981 & 96 & 889 & 10 & 975 & 4 & 981 & 3 & 982 \cr
\bottomrule
\end{tabular*}
\end{table*}

\begin{table}[width=\linewidth,cols=5]
\caption{Precision, Recall, and F1-Score of ML and Transformer-based Models for Balanced Dataset}
\label{tbl6}
\begin{tabular}{lcccc}
\toprule
Model & Class & Precision & Recall & F1-score \\
\midrule
\multirow{2}{*}{XGB} & ham & 0.98 & 0.99 & 0.98 \\
                     & spam & 0.99 & 0.98 & 0.98 \\
\midrule
\multirow{2}{*}{SVM} & ham & 1.00 & 1.00 & 1.00 \\
                     & spam & 1.00 & 1.00 & 1.00 \\
\midrule
\multirow{2}{*}{KNN} & ham & 0.91 & 1.00 & 0.95 \\
                     & spam & 1.00 & 0.90 & 0.95 \\
\midrule
\multirow{2}{*}{RF} & ham & 0.99 & 1.00 & 0.99 \\
                     & spam & 1.00 & 0.99 & 0.99 \\
\midrule
\multirow{2}{*}{DistilBERT} & ham & 1.00 & 1.00 & 1.00 \\
                            & spam & 1.00 & 1.00 & 1.00 \\
\midrule
\multirow{2}{*}{RoBERTa} & ham & 1.00 & 1.00 & 1.00 \\
                         & spam & 1.00 & 1.00 & 1.00 \\
\bottomrule
\end{tabular}
\end{table}

\begin{table}[width=\linewidth,cols=5, h]
\caption{Comparative Performance Analysis of Fine-tuned RoBERTa Model with Other Models on Balanced Data.}\label{tbl7}
\begin{tabular*}{\tblwidth}{@{} p{2cm}p{1cm}p{1.5cm}p{1cm}p{1.5cm}@{} }
\toprule
Model & Training Loss  & Training Accuracy & Testing Loss & Testing Accuracy\\
\midrule
XGB& 0.0325 & 99.72 & 0.0621 & 98.34\\
\midrule
SVM& 0.0041 & 99.86 & 0.0323 & 99.59\\
\midrule
KNN& 0.0457 & 97.76 & 0.1371 & 94.97\\
\midrule
RF& 0.0167 & 100 & 0.0530 & 99.38\\
\midrule
DistilBERT& 0.0015 & 99.97 & 0.0150 & 99.69\\
\midrule
RoBERTa& 0.0035 & 99.95 & 0.0122 & 99.84\\
\bottomrule
\end{tabular*}
\end{table}

We have compared our SMS spam detection study with literature studies that worked with similar elements in the same field of cybersecurity. The comparison, summarized in Table \ref{tbl8}, evaluates our work from several angles, including the datasets used, the machine and deep learning models used, and whether the study offers decision explainability and human interpretation for the algorithms.

\begin{table*}[htbp]
    \caption{Comparison of Our Study with Various Different Studies on SMS Spam Detection}\label{tbl8}
    \begin{tabularx}{\textwidth}{@{}lp{4cm}l>{\centering\arraybackslash}Xc@{}}
        \toprule
        Reference & Datasets & Models & Accuracy(\%) & Explainability Analysis \\
        \midrule
        \cite{asaju2021short} & UCI SMS spam collection & Naive Bayes & 99.42 & No\\
        \cite{xia2020discrete} & UCI SMS spam collection & HMM & 95.9 & No\\
        \cite{roy2020deep} & UCI SMS spam collection & CNN & 99.44 & No\\
        \cite{hossain2022detecting} & UCI SMS spam collection & BERT & 98.80 & No\\
        \cite{oswald2022spotspam} & UCI SMS spam collection & DistilBERT+SVM & 98.07 & No\\
        \textbf{This Study} & \textbf{UCI SMS spam collection} & \textbf{Fine-tuned RoBERTa} & \textbf{99.84} & \textbf{Yes} \\
        \bottomrule
    \end{tabularx}
\end{table*}

Our transparent research makes it stand out in the field of SMS spam detection. We have provided a detailed explanation of our suggested strategy, which performed better than the existing research works. Interestingly, none of the publications in Table \ref{tbl8} go as deep into a thorough human interpretation of the classifier's output as we have. They do not clarify the model's operation in spam detection.

\section{Explainability Analysis}

This study employs LIME and Transformers Interpret techniques to explain the predictions made by an improved RoBERTa model being used for spam message detection. These tools approximate the model's behavior by analyzing each word's contribution in the SMS text toward the final prediction. LIME and Transformers Interpret aim to answer “why the model made a particular prediction?”.  They achieve this by creating word attributions, which give each word in the text a positive or negative coefficient. This helps determine whether terms are more important for categorizing an SMS as "spam" or "ham" by showing the impact of each word on the model's prediction.  Positive coefficients indicate that the presence of the word contributes towards predicting a certain class, while negative coefficients indicate the opposite. The explainable model highlights words that scored positively or negatively for the prediction, which uses this information to visualize the full text. This visual aid facilitates comprehension of the rationale underlying the model's decision-making process by illuminating key terms that contributed significantly to the categorization result.

\begin{figure*}
    \begin{subfigure}{\textwidth}
        \centering
        \begin{subfigure}{0.4\textwidth}
            \centering
            \includegraphics[width=\linewidth, height=0.3\textwidth]{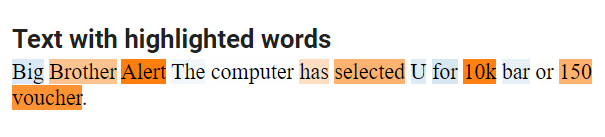}
            \caption{SMS 1st Text}
        \end{subfigure}
        \hfill
        \begin{subfigure}{0.5\textwidth}
            \centering
            \includegraphics[width=\linewidth, height=0.6\textwidth]{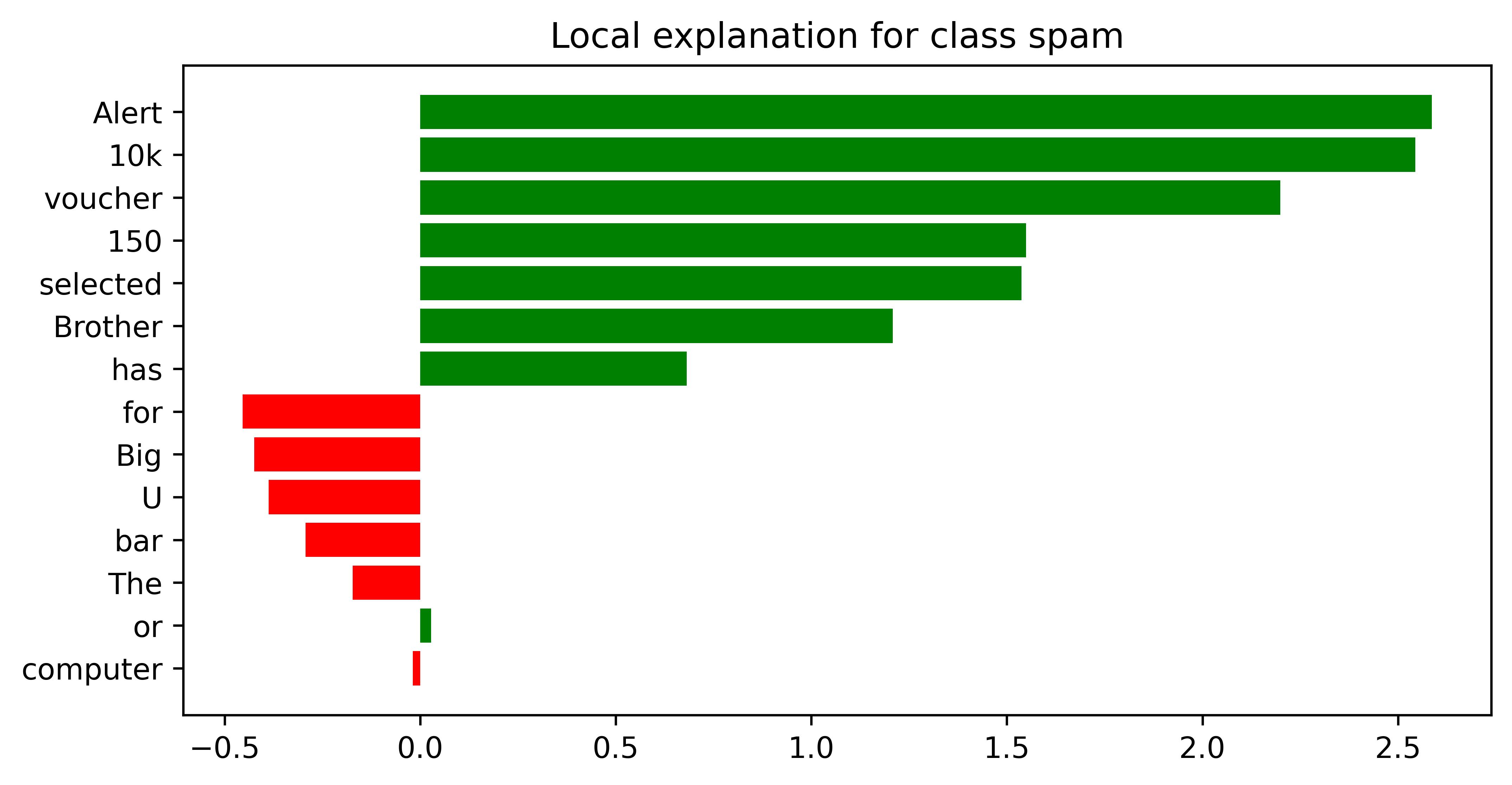}
            \caption{Explanation for the first text}
        \end{subfigure}
    \end{subfigure}

    \begin{subfigure}{\textwidth}
        \centering
        \begin{subfigure}{0.4\textwidth}
            \centering
            \includegraphics[width=\linewidth, height=0.3\textwidth]{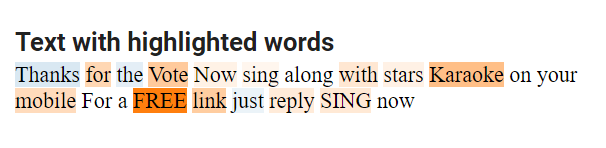}
            \caption{SMS 2nd Text}
        \end{subfigure}
        \hfill
        \begin{subfigure}{0.5\textwidth}
            \centering
            \includegraphics[width=\linewidth, height=0.6\textwidth]{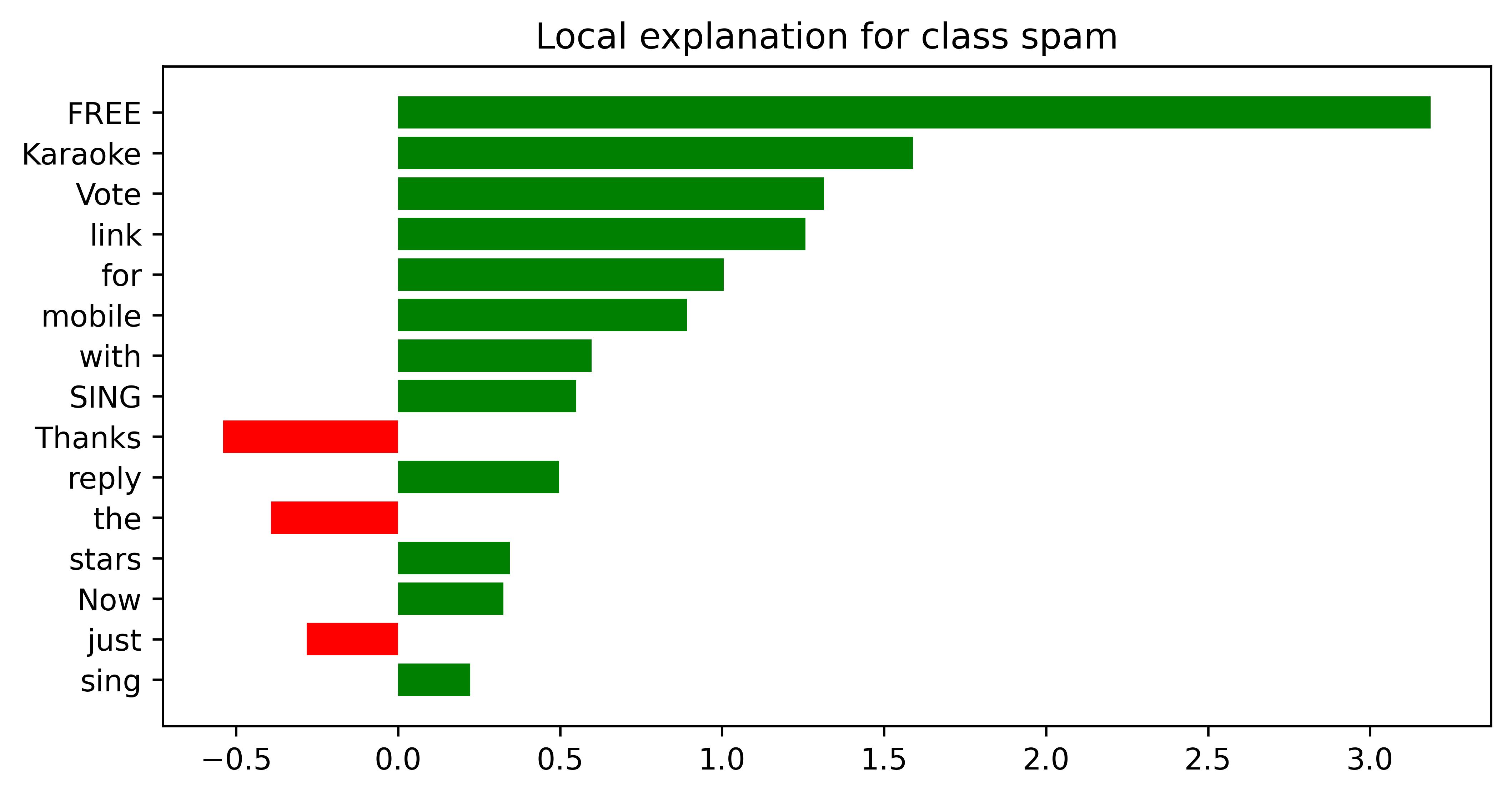}
            \caption{Explanation for the second text}
        \end{subfigure}
    \end{subfigure}

    \caption{LIME Explanation for SMS Spam Texts}
    \label{fig:LIME-Explanation-Spam}
\end{figure*}

\begin{figure*}
    \begin{subfigure}{\textwidth}
        \centering
        \begin{subfigure}{0.4\textwidth}
            \centering
            \includegraphics[width=\linewidth, height=0.3\textwidth]{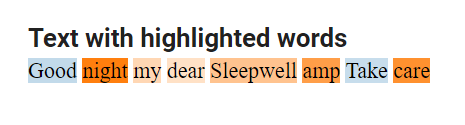}
            \caption{SMS 3rd Text}
        \end{subfigure}
        \hfill
        \begin{subfigure}{0.5\textwidth}
            \centering
            \includegraphics[width=\linewidth, height=0.6\textwidth]{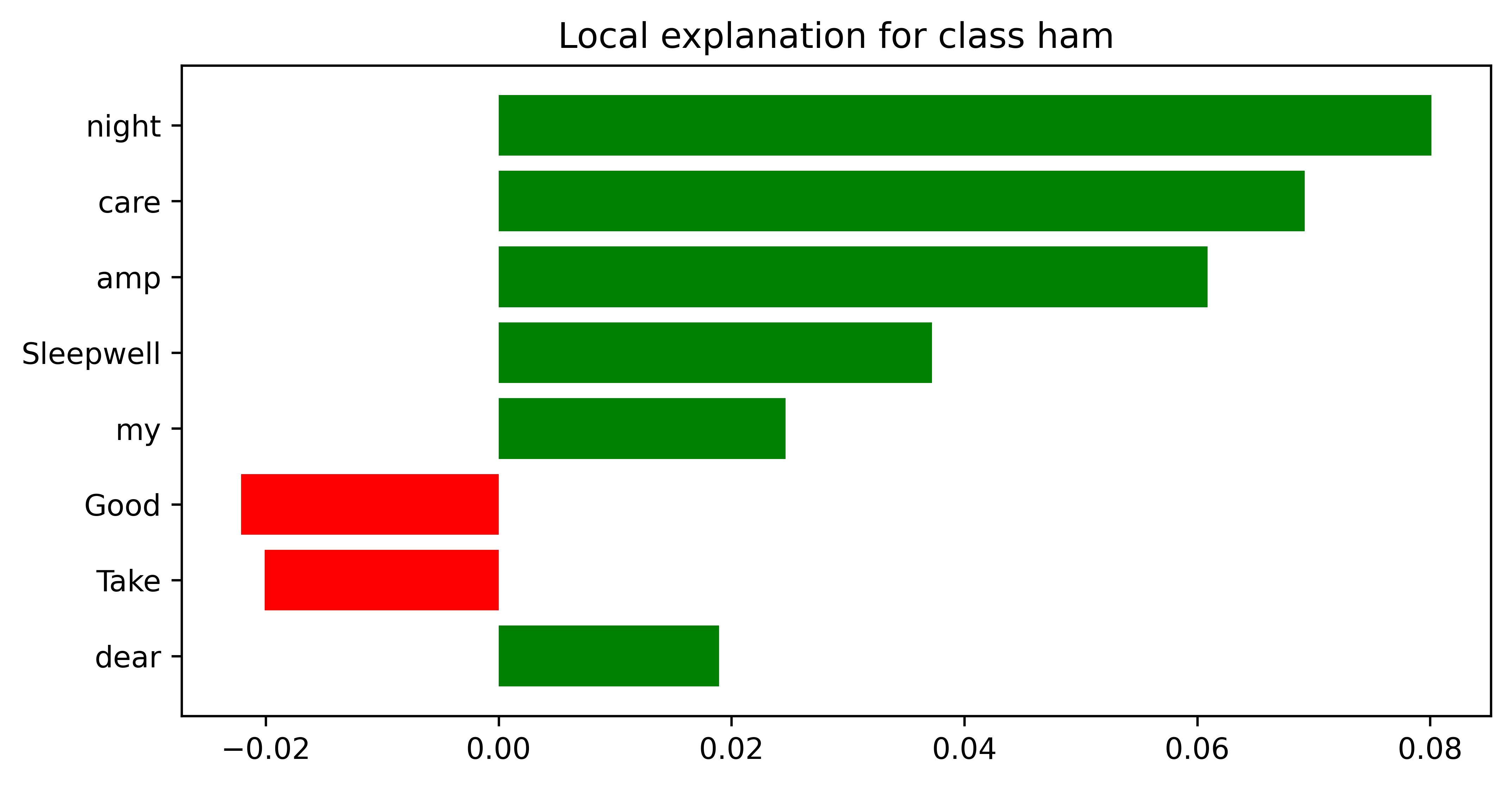}
            \caption{Explanation for the Third text}
        \end{subfigure}
    \end{subfigure}

    \begin{subfigure}{\textwidth}
        \centering
        \begin{subfigure}{0.4\textwidth}
            \centering 
            \includegraphics[width=\linewidth, height=0.3\textwidth]{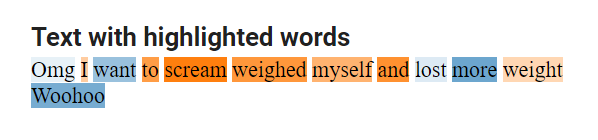}
            \caption{SMS 4th Text}
        \end{subfigure}
        \hfill
        \begin{subfigure}{0.5\textwidth}
            \centering
            \includegraphics[width=\linewidth, height=0.6\textwidth]{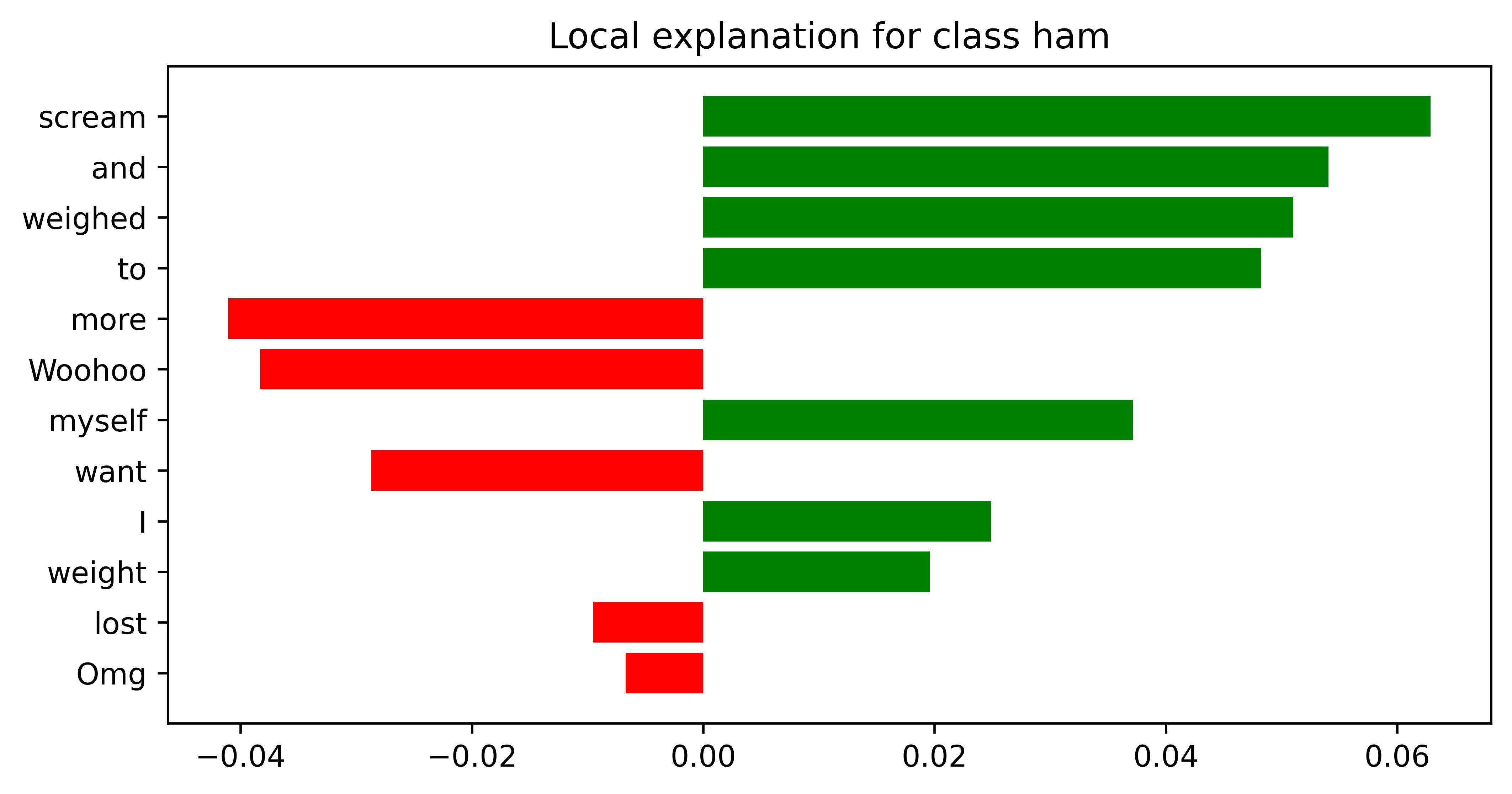}
            \caption{Explanation for the Fourth text}
        \end{subfigure}
    \end{subfigure}

    \caption{LIME Explanation for SMS Ham Texts}
      \label{fig:LIME-Explanation-Ham}
\end{figure*}

In LIME, the corresponding predictions from the complicated model are fitted alongside an interpretable model to the perturbed samples. This procedure is essential to the LIME process to produce better interpretable explanations of the complex model's predictions. This interpretable model's coefficients show how much each feature (word) contributed to creating the predictions. Sometimes, LIME will normalize the data to ensure that the importance rankings are consistent across different models or explanations. In an effort to reduce dimensionality, the study set the initial number of features (words) in the LIME explainer to 15 to accelerate the computational explanation process. Limiting the characteristics simplifies the computational explanation process and provides a clearer, more manageable list of words that have a major influence on the model's prediction for a particular text instance. Fig. \ref{fig:LIME-Explanation-Spam} and Fig. \ref{fig:LIME-Explanation-Ham} illustrate the explainability of the RoBERTa model using LIME for both "spam" and "ham", while Table \ref{tbl9}, Table \ref{tbl10}, Table \ref{tbl11} and Table \ref{tbl12} presents the positive and negative coefficients associated with terms to aid users in understanding spam detection.

\begin{table*}[htbp]
\caption{The positive and negative coefficients using LIME for the first text}\label{tbl9}
\begin{tabular*}{\textwidth}{@{\extracolsep{\fill}}cccc}
\toprule
\multicolumn{2}{c}{Positive coefficients} & \multicolumn{2}{c}{Negative coefficients} \\
\cmidrule{1-2}\cmidrule{3-4}
Word & Value & Word & Value \\
\midrule
`Alert' & 2.586283490431408 & `for' & -0.4540554753154869 \\
`10k' & 2.5443085492464883 & `Big' & -0.424707563134796 \\
`voucher' & 2.19896727844678 & `U' & -0.3875706185927502 \\
`150' & 1.5489612217746545 & `bar' & -0.2933243630343485 \\
`selected' & 1.5372338239788719 & `The' & -0.17252983760850746 \\
`Brother' & 1.2084815552598014 & `computer' & -0.019025118002313234  \\
`has' & 0.6819267212480585 &  &  \\
`or' & 0.02814567596003713 &  &  \\
\bottomrule
\end{tabular*}
\end{table*}

\begin{table*}[htbp]
\caption{The positive and negative coefficients using LIME for the second text}\label{tbl10}
\begin{tabular*}{\textwidth}{@{\extracolsep{\fill}}cccc}
\toprule
\multicolumn{2}{c}{Positive coefficients} & \multicolumn{2}{c}{Negative coefficients} \\
\cmidrule{1-2}\cmidrule{3-4}
Word & Value & Word & Value \\
\midrule
`FREE' & 3.186586145122694 & `Thanks' & -0.5401725013063734 \\
`Karaoke' & 1.5886755775421426 & `the' & -0.3920060404493261 \\
`Vote' & 1.314393037077685 & `just' & -0.28230176631326165 \\
`link' & 1.257098816616926 &  &  \\
`for' & 1.0052090127637938 &  &  \\
`mobile' & 0.8919310037119642 &  &  \\
`with' & 0.597081408704299 &  &  \\
`SING' & 0.5501241601784591 &  &  \\
`reply' & 0.49702113599413067 &  &  \\
`stars' & 0.3446891653595809 &  &  \\
`Now' & 0.3248660570258343 &  &  \\
`sing' & 0.2221033541586965 &  &  \\
\bottomrule
\end{tabular*}
\end{table*}

\begin{table*}[htbp]
\caption{The positive and negative coefficients using LIME for the third text}\label{tbl11}
\begin{tabular*}{\textwidth}{@{\extracolsep{\fill}}cccc}
\toprule
\multicolumn{2}{c}{Positive coefficients} & \multicolumn{2}{c}{Negative coefficients} \\
\cmidrule{1-2}\cmidrule{3-4}
Word & Value & Word & Value \\
\midrule
`night' & 0.080110152925621 & `Good' & -0.022107989126383402 \\
`care' & 0.06923821968423888 & `Take' & -0.020104893561268274 \\
`amp' & 0.060886333137515174 &  &  \\
`Sleepwell' & 0.03722758230659266 &  &  \\
`my' & 0.024641469910968265 &  &  \\
`dear' & 0.018922945727309903 &  &  \\
\bottomrule
\end{tabular*}
\end{table*}

\begin{table*}[htbp]
\caption{The positive and negative coefficients using LIME for the fourth text}\label{tbl12}
\begin{tabular*}{\textwidth}{@{\extracolsep{\fill}}cccc}
\toprule
\multicolumn{2}{c}{Positive coefficients} & \multicolumn{2}{c}{Negative coefficients} \\
\cmidrule{1-2}\cmidrule{3-4}
Word & Value & Word & Value \\
\midrule
`scream' & 0.0628967823689254 & `more' & -0.04109442935123623 \\
`and' & 0.05404817757012105 & `Woohoo' & -0.03831983060032088 \\
`weighed' & 0.05100704096103593 & `want' & -0.028705541788966896 \\
`to' & 0.04824724743333108 & `lost' & -0.009499243825646576 \\
`myself' & 0.037156130726796105 & `Omg' & -0.006713896998241079 \\
`I' & 0.02486380111242408 &  &  \\
`weight' & 0.019592668979617558 &  &  \\
\bottomrule
\end{tabular*}
\end{table*}

\begin{figure*}[htbp]
    \centering
    \begin{subfigure}{\textwidth}
        \centering
        \includegraphics[width=0.9\textwidth]{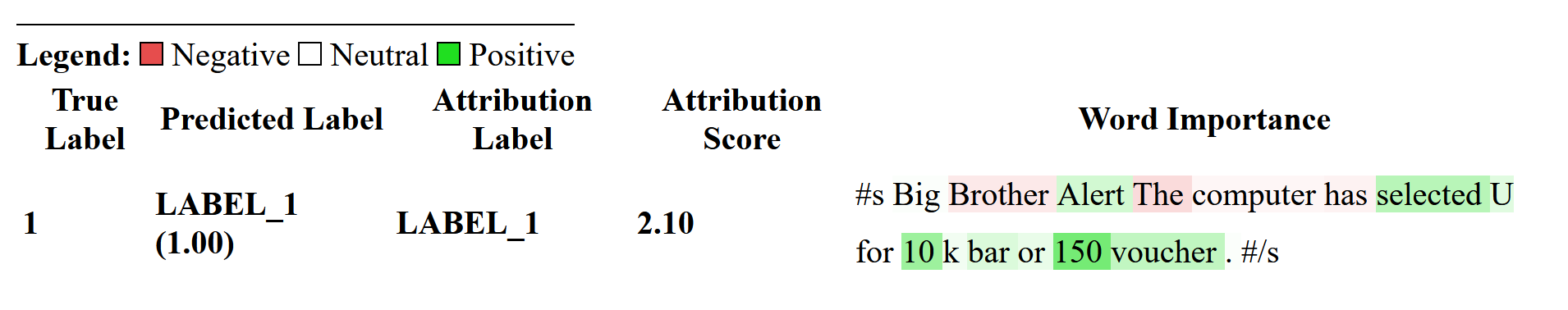}
        \caption{Explanation for first text (spam)}
    \end{subfigure}

        \begin{subfigure}{\textwidth}
        \centering
        \includegraphics[width=0.9\textwidth]{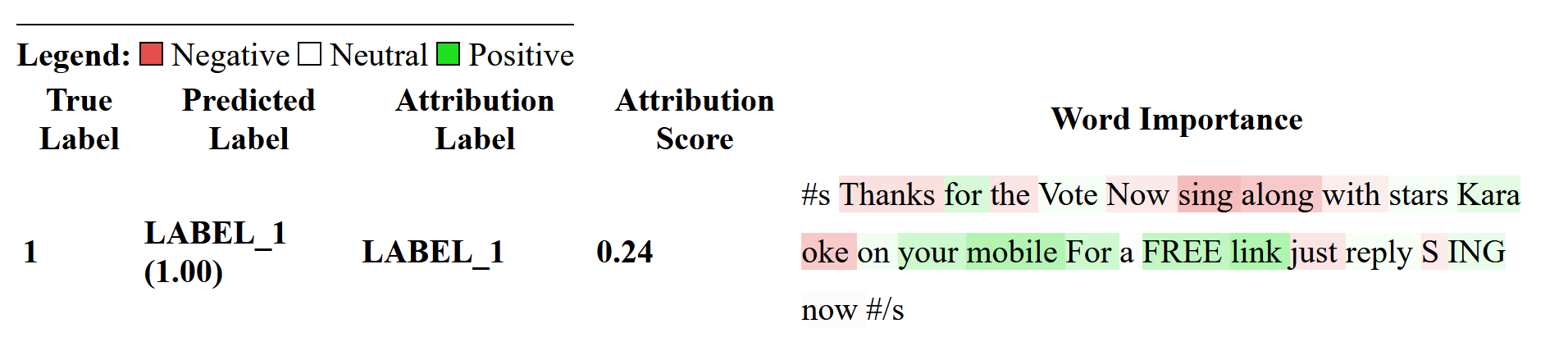}
        \caption{Explanation for second text (spam)}
    \end{subfigure}
    
    \caption{Transformers Interpret Explanation for two Spam SMS Text}\label{fig:Interpret_Explanation-spam}
\end{figure*}

\begin{figure*}[htbp]
    \centering
            \begin{subfigure}{\textwidth}
        \centering
        \includegraphics[width=0.9\textwidth]{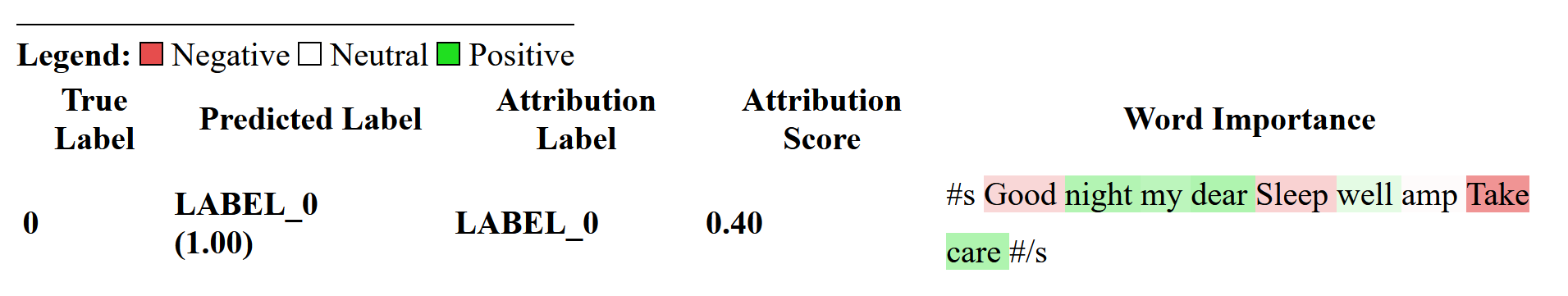}
        \caption{Explanation for Third text (ham)}
    \end{subfigure}

            \begin{subfigure}{\textwidth}
        \centering
        \includegraphics[width=0.9\textwidth]{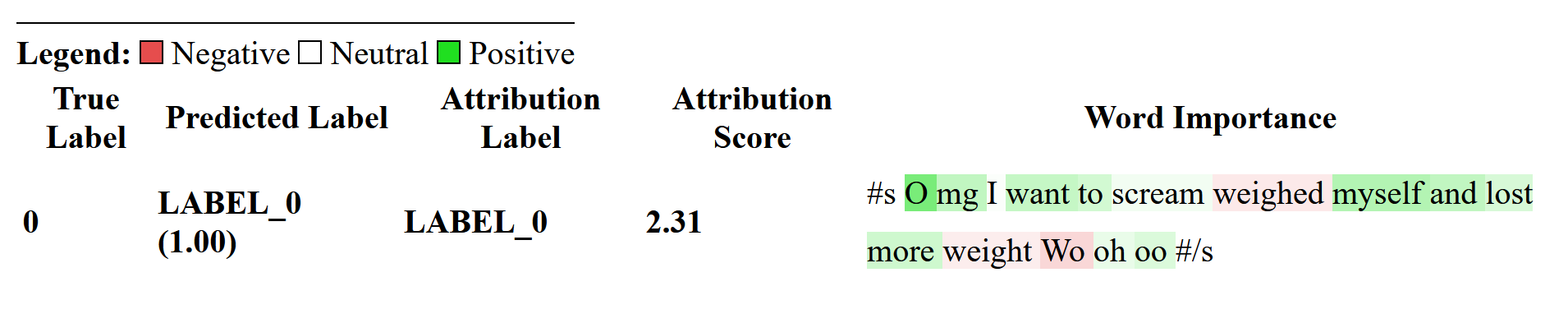}
        \caption{Explanation for Fourth text (ham)}
    \end{subfigure}
    
    \caption{Transformers Interpret Explanation for two ham SMS Text}\label{fig:Interpret_Explanation-ham}
\end{figure*}

We used Transformers Interpret in addition to LIME to assess the model interpretability with similar SMS texts. This Transformers Interpret illustrates how the transformer-based RoBERTa model uses the tokenizer to divide the huge words into subwords with the prefix "\#\#". This occurs when the model encounters complex or out-of-vocabulary words, which it then breaks down into smaller sub-words to simplify processing. The model can manage a broader vocabulary and comprehend the context of complicated words more accurately due to this procedure. The Transformers Interpret tool's interpretation findings for SMS spam messages are shown in Fig. \ref{fig:Interpret_Explanation-spam} and Fig. \ref{fig:Interpret_Explanation-ham}, and Table \ref{tbl13}, Table \ref{tbl14}, Table \ref{tbl15} and Table \ref{tbl16} show the relevance of the words in four SMS, with positive and negative coefficients showing each word's influence on the model's prediction. In this experiment, we worked with a binary classification task where "True Label" = 0 denotes "ham" and "True Label" = 1 denotes "spam".

\begin{table*}[htbp]
\caption{The positive and negative coefficients using Transformer Interpret for the first text}\label{tbl13}
\begin{tabular*}{\textwidth}{@{\extracolsep{\fill}}cccc}
\toprule
\multicolumn{2}{c}{Positive coefficients} & \multicolumn{2}{c}{Negative coefficients} \\
\cmidrule{1-2}\cmidrule{3-4}
Word & Value & Word & Value \\
\midrule
`Big' & 0.024297947664629425 & `Brother' & -0.14376454412860262 \\
`Alert' & 0.20062465985674183 & `The' & -0.20299253710630974 \\
`selected' & 0.3235594150691528 & `computer' & -0.06358605308176257 \\
`U' & 0.14328277724673683 & `has' & -0.09203341078407345 \\
`10' & 0.454875630595537 & `for' & -0.013336465891418426 \\
`k' & 0.07143031221542949 &  &  \\
`bar' & 0.16807094091950164 & & \\
`150' & 0.6246831675560521 & & \\
`or' & 0.11691662255361639 &  &  \\
`voucher' & 0.29147154979015566 & & \\
`.' & 0.023613104082805682 & & \\
`' & 0.17233550519021407 &  & \\
\bottomrule
\end{tabular*}
\end{table*}

\begin{table*}[htbp]
\caption{The positive and negative coefficients using Transformers Interpret for the second text}\label{tbl14}
\begin{tabular*}{\textwidth}{@{\extracolsep{\fill}}cccc}
\toprule
\multicolumn{2}{c}{Positive coefficients} & \multicolumn{2}{c}{Negative coefficients} \\
\cmidrule{1-2}\cmidrule{3-4}
Word & Value & Word & Value \\
\midrule
`for' & 0.18965947100466987 & `Thanks' & -0.18213375681044702 \\
`Vote' & 0.04214511554119887 & `the' & -0.16879623241117406 \\
`stars' & 0.050776681629629485 & `sing' & -0.3819342785384402 \\
`Kara' & 0.13869346357463316 & `along' & -0.32168706815694476 \\
`on' & 0.07859092995531514 & `with' & -0.10687877218596162 \\
`your' & 0.2284359424426384 & `just' & -0.15317418612821526 \\
`For' & 0.2290909616422906 & `S' & -0.12666676012232256 \\
`FREE' & 0.2880400601164391 & `now' & -0.031842205592812194 \\
`link' & 0.37679605294057267 &  & \\
`reply' & 0.04222149540549413 &  & \\
`ING' & 0.10210499982518297 &  & \\
`mobile' & 0.3437490059251579 &  & \\
`a' & 0.03278371376214245 &  & \\
\bottomrule
\end{tabular*}
\end{table*}

\begin{table*}[htbp]
\caption{The positive and negative coefficients using Transformers Interpret for the third text}\label{tbl15}
\begin{tabular*}{\textwidth}{@{\extracolsep{\fill}}cccc}
\toprule
\multicolumn{2}{c}{Positive coefficients} & \multicolumn{2}{c}{Negative coefficients} \\
\cmidrule{1-2}\cmidrule{3-4}
Word & Value & Word & Value \\
\midrule
`night' & 0.3506601184212354 & `Good' & -0.23430671367151 \\
`my' & 0.30588133708098775 & `Sleep' & -0.2558005048167494 \\
`dear' & 0.3750717679647341 & `Take' & -0.6100631898260275 \\
`care' & 0.36718414960882323 & `amp' & -0.025046901303540856 \\
 &  &  & \\
\bottomrule
\end{tabular*}
\end{table*}

\begin{table*}[htbp]
\caption{The positive and negative coefficients using Transformers Interpret for the fourth text}\label{tbl16}
\begin{tabular*}{\textwidth}{@{\extracolsep{\fill}}cccc}
\toprule
\multicolumn{2}{c}{Positive coefficients} & \multicolumn{2}{c}{Negative coefficients} \\
\cmidrule{1-2}\cmidrule{3-4}
Word & Value & Word & Value \\
\midrule
`O' & 0.6168087105189476 & `weighed' & -0.14799138790010427 \\
`mg' & 0.285592216107348 & `weight' & -0.1102382677942815 \\
`I' & 0.02889748690680515 & `Wo' & -0.23351373922826477 \\
`want' & 0.2613656158757424 &  & \\
`to' & 0.21709997128226066 &  & \\
`scream' & 0.0764171095678442 &  & \\
`myself' & 0.3498253922866739 &  & \\
`and' & 0.2869778663987718 &  & \\
`lost' & 0.18199395807032306 &  & \\
`more' & 0.22200118513344438 &  & \\
`oh' & 0.11642180604530727 &  & \\
`oo' & 0.1630583357991135 &  & \\
\bottomrule
\end{tabular*}
\end{table*}

\section{Discussion}

The overall experiment was conducted with both imbalanced and balanced datasets. The performance of the proposed fine-tuned RoBERTa model was measured with an imbalanced and balanced dataset. We mainly worked with imbalanced and balanced datasets to see our optimized and fine-tuned model’s effectiveness. We employed a data-balancing technique for this model’s effectiveness visualization and balanced the data. On the imbalanced dataset, the model achieved 99.69\% training and 99.37\% testing accuracy with 0.0292 loss. On the other hand,  on the balanced dataset, our fine-tuned model achieved 99.95\% training accuracy and 99.84\% testing accuracy with 0.0122 loss. The performance accuracy of this model using an imbalanced and balanced dataset does not show too much difference, just showing 0.47\% of difference between these two values. To compare with our model performance, we selected several traditional ML models and transformer-based models on both dataset types. Among those tested models, the RoBERTa model performed better. Furthermore, the optimized model performed well among those tested models and outperformed previous studies, demonstrating its effectiveness.

Along with the performance analysis of our model, we also worked with XAI techniques which can provide answers to various types of questions such as "Why did the model make a particular prediction?" and "Which features have the most significant impact on the model's output?" related to the functioning and decision-making process of the model. For this, we used two well-known XAI techniques such as LIME and Transformers Interpret to understand the models' decision-making process. It is more difficult to understand how RoBERTa model makes its predictions because, as we know, it is a transformer-based model which is so complex in architecture. Feature importance analysis using XAI, helps users to understand the factors that most influence a model's decision-making process. By analyzing input features such as words or sub-words, XAI methods provide insights into the factors that contributed significantly to the specific classification of spam or ham messages. Mainly, those techniques calculated each word or sub-word coefficient where some words indicate a positive coefficient and some negative coefficient which indicates that these words with positive coefficients are considered more indicative to classify a class, while these words with negative coefficients are considered more indicative to do not support the predicted class. This overall process of explanation shows the transparency of our proposed model by which users can gain insights into the prediction. Moreover, this experimental analysis only used one benchmark dataset of spam messages. In the future, we will collect other different datasets of SMS spam to explore our fine-tuned model performance and its explainability on new data.

\section{Conclusion}

Utilizing cutting-edge natural language processing algorithms has significantly improved SMS spam detection, demonstrating the efficacy of these methods. Using LLMs has enormous potential to improve the global computing user experience and address long-standing societal issues. In this paper, we utilized modified transformer-based LLMs (DistilBERT and RoBERTa) to identify SMS spam. A benchmark dataset was utilized for the entire investigation, although data imbalance issues afflicted it. However, these challenges were effectively mitigated through the implementation of data augmentation such as back translation. The experimental results showed that, compared to traditional ML models, our transformer-based model, more specifically RoBERTa performed better in differentiating between spam and valid SMSs. To achieve such high performance, the model is primarily optimized and fine-tuned with crucial hyperparameters defined, and trained on a particular subset of real data. This improved version of RoBERTa achieved a testing accuracy of 99.84\% in the SMS spam classification task. Moreover, explainable AI techniques such as LIME and Transformers Interpret are indeed used to see the explanations of the model prediction by calculating the word coefficient. By using the explanation, people can learn how the model works and why it makes its decisions. This increases user understanding and confidence in the model's results through transparency, strengthening the model's overall robustness and reliability.

The experimental results demonstrate an enhancement of our proposed transformer-based model compared to previously proposed approaches of others in SMS spam detection. LLMs in particular are being adopted quickly, which is driving ongoing innovation and growth in the field of natural language processing. In the future, our research aims to expand our spam detection work by exploring a diverse range of models and datasets. Furthermore, analyzing models in-depth to strengthen their robustness using explainable AI approaches is consistent with the continuous endeavors to raise the trustworthiness and understanding of AI systems.

\printcredits

\bibliographystyle{model1-num-names}

\bibliography{cas-refs}


\end{document}